\documentclass{article} % For LaTeX2e
\usepackage{iclr2026_conference,times}
\usepackage{multirow}
\usepackage{xcolor}

\usepackage{amsmath,amsfonts,bm}

\def\eqref#1{equation~\ref{#1}}
\def\1{\bm{1}}

\DeclareMathAlphabet{\mathsfit}{\encodingdefault}{\sfdefault}{m}{sl}
\SetMathAlphabet{\mathsfit}{bold}{\encodingdefault}{\sfdefault}{bx}{n}

\usepackage{url}
\usepackage{hyperref}
\usepackage{times}
\usepackage{graphicx}
\usepackage{algorithm}
\usepackage{wrapfig}
\usepackage{algorithmic}
\usepackage{subcaption}
\usepackage{tcolorbox}
\usepackage[normalem]{ulem}
\useunder{\uline}{\ul}{}
\usepackage{booktabs}
\usepackage[table,xcdraw]{xcolor}
\definecolor{myYellow}{RGB}{240,173,0}
\definecolor{myGreen}{RGB}{0,158,115}
\definecolor{myBlue}{RGB}{0,114,178}
\definecolor{myPurple}{RGB}{170,51,119}

\usepackage{graphicx} % Required for inserting images
\usepackage{amsmath, amssymb, amsthm}
\usepackage{hyperref}
\usepackage{enumitem}

\newtheorem{theorem}{Theorem}[section]
\newtheorem{definition}[theorem]{Definition}
\newtheorem{assumption}[theorem]{Assumption}
\newtheorem{lemma}[theorem]{Lemma}

\newtheorem*{remark}{Remark}

\DeclareMathOperator{\ent}{Ent} % Entropy
\newcommand{\prob}{\mathbb{P}} % Probability

\newcommand{\res}{\mathcal{R}} % Observed Response
\newcommand{\imp}{\mathcal{I}} % Implicit Thought
\newcommand{\process}{R_\text{process}^{(t)}} % Process Reward
\newcommand{\total}{R_\text{total}} % Total Reward

\title{Information Gain-based Policy Optimization: A Simple and Effective Approach for Multi-Turn Search Agents}

\author{Guoqing Wang$^{1}$\thanks{Equal contributions.},
Sunhao Dai$^{1}$\textsuperscript{*}\thanks{Corresponding author.},
Guangze Ye$^{3}$\textsuperscript{*}, Zeyu Gan$^{2}$, Wei Yao$^{2}$, Yong Deng$^{1}$\\
\textbf{Xiaofeng Wu$^{1}$, Zhenzhe Ying$^{1}$}\\
$^{1}$Venus Team, Ant Group, 
$^{2}$Renmin University of China, 
$^{3}$Individual Author\\
\texttt{\{guoqingwang905, sunhaodai, guangzeye98\}@gmail.com},\\
\texttt{zygan@ruc.edu.cn, wei.yao@ruc.edu.cn},\\
\texttt{\{dengyong.deng, congyu.wxf, zhenzhe.yzz\}@antgroup.com}
}

\iclrfinalcopy % Uncomment for camera-ready version, but NOT for submission.
\begin{document}

\maketitle

\begin{abstract}
Large language model (LLM)-based agents are increasingly trained with reinforcement learning (RL) to enhance their ability to interact with external environments through tool use, particularly in search-based settings that require multi-turn reasoning and knowledge acquisition. However, existing approaches typically rely on outcome-based rewards that are only provided exclusively upon generating the final answer. This reward sparsity becomes particularly problematic in multi-turn settings, where long trajectories exacerbate three critical issues: (i) advantage collapse, where all rollouts receive identical rewards and provide no useful learning signals; (ii) lack of fine-grained credit assignment, where the correctness of intermediate turns is obscured, especially in long-horizon tasks; and (iii) poor sample efficiency, where each rollout yields only a single outcome signal, leading to low data utilization. In this paper, we propose Information Gain-based Policy Optimization (IGPO), a simple yet effective RL framework that provides dense and intrinsic supervision for multi-turn agent training. IGPO models each interaction turn as an incremental process of acquiring information about the ground truth, and defines turn-level rewards as the marginal increase in the policy's probability of producing the correct answer. 
Unlike prior process-level reward approaches that depend on external reward models or costly Monte Carlo estimation, IGPO derives intrinsic rewards directly from the model's own belief updates.
These intrinsic turn-level rewards are combined with outcome-level supervision to form dense reward signals. Extensive experiments on both in-domain and out-of-domain benchmarks demonstrate that IGPO consistently outperforms strong baselines in multi-turn scenarios, achieving higher accuracy and improved data efficiency. Our code is available at \url{https://github.com/GuoqingWang1/IGPO}.
\end{abstract}

\section{Introduction}

Large language model (LLM)–based agents are increasingly equipped with the ability to interact with external environments through tool use~\citep{zhang2025landscape, huang2025deep, li2025torl}, a capability often regarded as a critical step toward building general-purpose autonomous intelligent systems~\citep{gutierrez2023proposal, qu2025tool}. One of the most prominent application scenarios is \emph{agentic search}, where an agent invokes tools such as web search~\citep{zhang2025web, qi2024webrl} to access up-to-date, large-scale knowledge that substantially enhances its ability to solve complex, knowledge-intensive tasks~\citep{ning2025survey}.
Through iterative interaction with the external environment via such tools, search agents can gradually acquire missing information and refine their reasoning trajectories toward solving the target query.

To equip general-purpose LLMs with such agentic search capabilities, early efforts primarily relied on prompt-based workflows~\citep{li2025search, wang2024searching, zheng2024openresearcher}, which allowed tool use without additional training but often suffered from poor generalization. More recent studies have explored supervised fine-tuning (SFT)~\citep{wang2024corag} and reinforcement learning (RL)~\citep{jin2025search, song2025r1, zheng2025deepresearcher} to explicitly incentivize tool use, achieving markedly better performance. In particular, Group Relative Policy Optimization (GRPO)~\citep{shao2024deepseekmath}–style methods have emerged as the dominant approach for training agentic LLMs. In this paradigm, a group of rollouts is generated for each query under the current policy, and outcome-based rewards, typically defined by the correctness of the final answer against the ground truth, are used to construct group-relative advantages that drive policy optimization.

Despite their simplicity and effectiveness in relatively easy tasks, outcome rewards suffer from inherent sparsity~\citep{zhang2025process}, providing supervision exclusively at the final answer. This limitation is particularly detrimental in multi-turn agentic settings, where long trajectories exacerbate three critical issues.
\textbf{First}, outcome-only rewards frequently lead to \emph{advantage collapse}: when intra-group rollouts yield identical answers (e.g., uniformly correct or incorrect), they receive identical rewards, yielding zero group-relative advantages and no gradient signal.
\begin{wrapfigure}{r}{0.42\textwidth} % r表示靠右，l表示靠左
    \vspace{-5pt}
    \centering
    \includegraphics[width=0.40\textwidth]{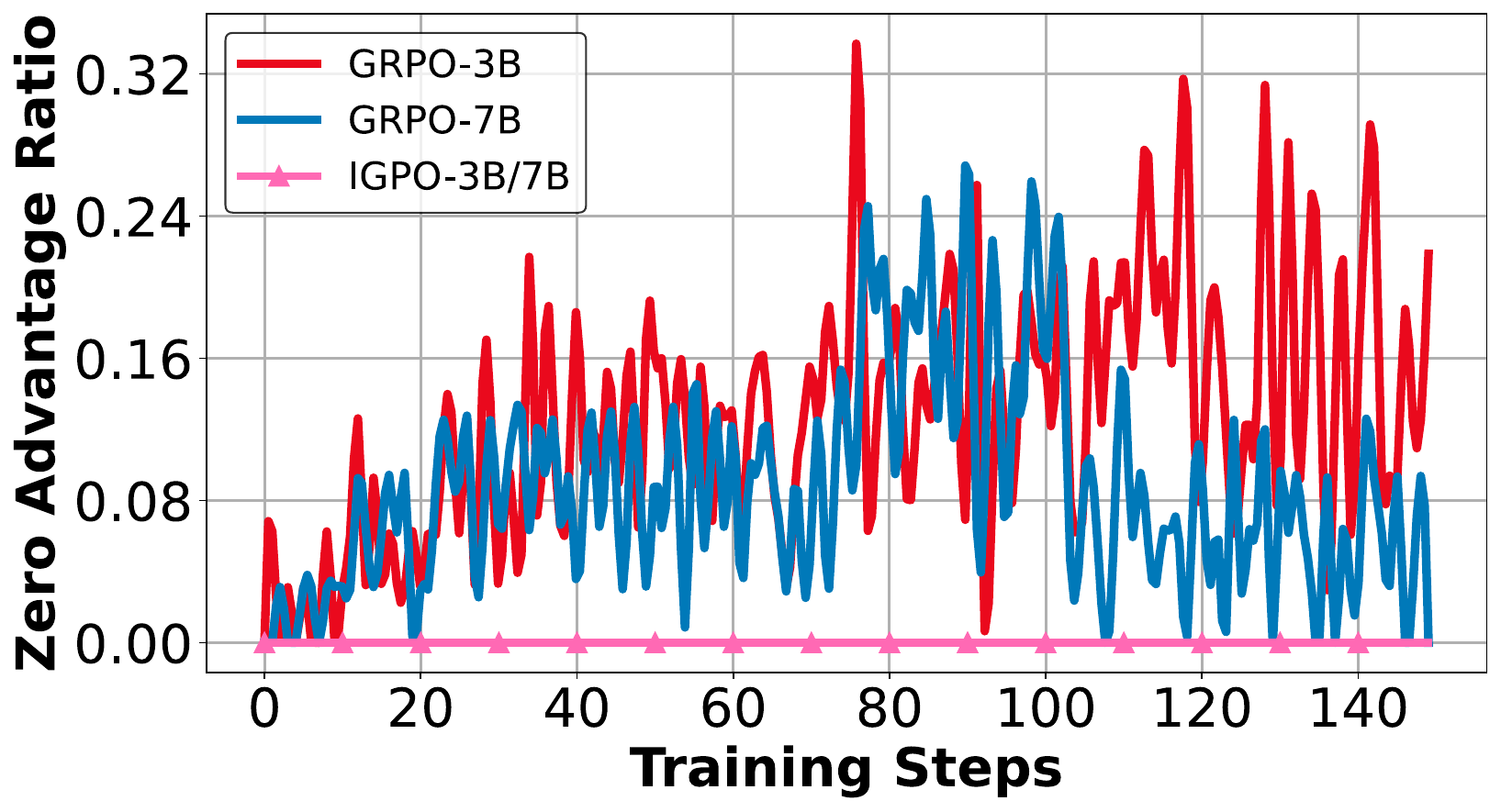}
    \vspace{-10pt}
    \caption{Proportion of zero-advantage groups during training—IGPO vs. GRPO on Qwen2.5-3B/7B-Instruct.}
    \vspace{-15pt}
    \label{fig:adv_collapse}
\end{wrapfigure}
As shown in \autoref{fig:adv_collapse}, a substantial portion of training iterations suffer from this issue, especially for smaller models, which struggle more with complex queries. \textbf{Second}, outcome supervision \emph{fails to provide fine-grained credit assignment}. In multi-turn scenarios, later turns are tightly dependent on earlier ones: the action of the current turn may be correct but rendered useless by prior mistakes, or conversely, early successes may be negated by subsequent errors. Thus, distinguishing the correctness of intermediate turns is essential in this scenario, but outcome rewards tend to blur such specific correctness. \textbf{Third}, outcome reward sparsity results in \emph{poor sample efficiency}. Since the entire trajectory receives only a single terminal signal, dense intermediate information is wasted, requiring significantly more samples to learn effective policies.

Recent approaches have attempted to mitigate these issues by introducing process rewards. One line of work leverages external oracle knowledge or reward models to judge intermediate steps~\citep{wang2025stepsearch, feng2025group}, but this strategy is costly and risks introducing bias. Another line relies on Monte Carlo simulations to estimate step values~\citep{wang2023math, zuo2025ttrl, zhang2025process}, yet these methods suffer from high variance unless a large number of samples are collected. Overall, both directions face challenges in scalability and fail to provide simple and stable supervision, underscoring the need for an intrinsic and reliable process reward design.

To address these challenges, we propose Information-Gain-based Policy Optimization (IGPO), a simple yet effective RL framework that provides stable and intrinsic supervision for multi-turn search agent training. The key intuition is to model each agent–environment interaction turn as an incremental process of acquiring information about the ground truth. Specifically, at every turn, IGPO computes the policy’s probability of producing the correct answer and defines the turn-level reward as the marginal increase in this probability compared to the previous state. This information gain reward offers ground-truth-aware feedback at every turn, in contrast to outcome rewards that only supervise the final answer. 
However, because outcome rewards are still necessary to anchor learning to the final objective, IGPO incorporates them alongside the turn-level rewards to construct dense reward signals for each rollout.
To further stabilize training, we normalize the information gain rewards and outcome rewards separately within groups and propagate them with discounted accumulation, enabling the computation of turn-level discounted returns that capture long-horizon dependencies. Finally, IGPO optimizes the policy with a GRPO-style surrogate objective, replacing rollout-level advantages with our turn-level discounted returns. Additionally, we introduce a vectorized implementation to minimize the computational overhead of information gain rewards.

To evaluate the effectiveness of IGPO, we conduct extensive experiments on both in-domain and out-of-domain benchmarks with search-based agents. Results show that IGPO consistently outperforms strong baselines, delivering substantial gains in both answer accuracy and sample efficiency.
Our main contributions can be summarized as follows: 
(1) We analyze the phenomenon of advantage collapse in outcome reward–based optimization, and reveal the inefficiency of existing process-level rewards due to reliance on external knowledge or high-variance estimation.
(2) We propose IGPO, a simple yet effective policy optimization framework that leverages turn-level information gain to provide dense, ground-truth-aware supervision with negligible computational overhead. (3) Comprehensive experiments demonstrate that IGPO outperforms strong baselines across multiple benchmarks and significantly improves sample efficiency, especially for smaller models.

\section{Preliminaries}

\subsection{Task Formulation}
\begin{figure}[t] 
    \centering
    \includegraphics[width=1\textwidth]{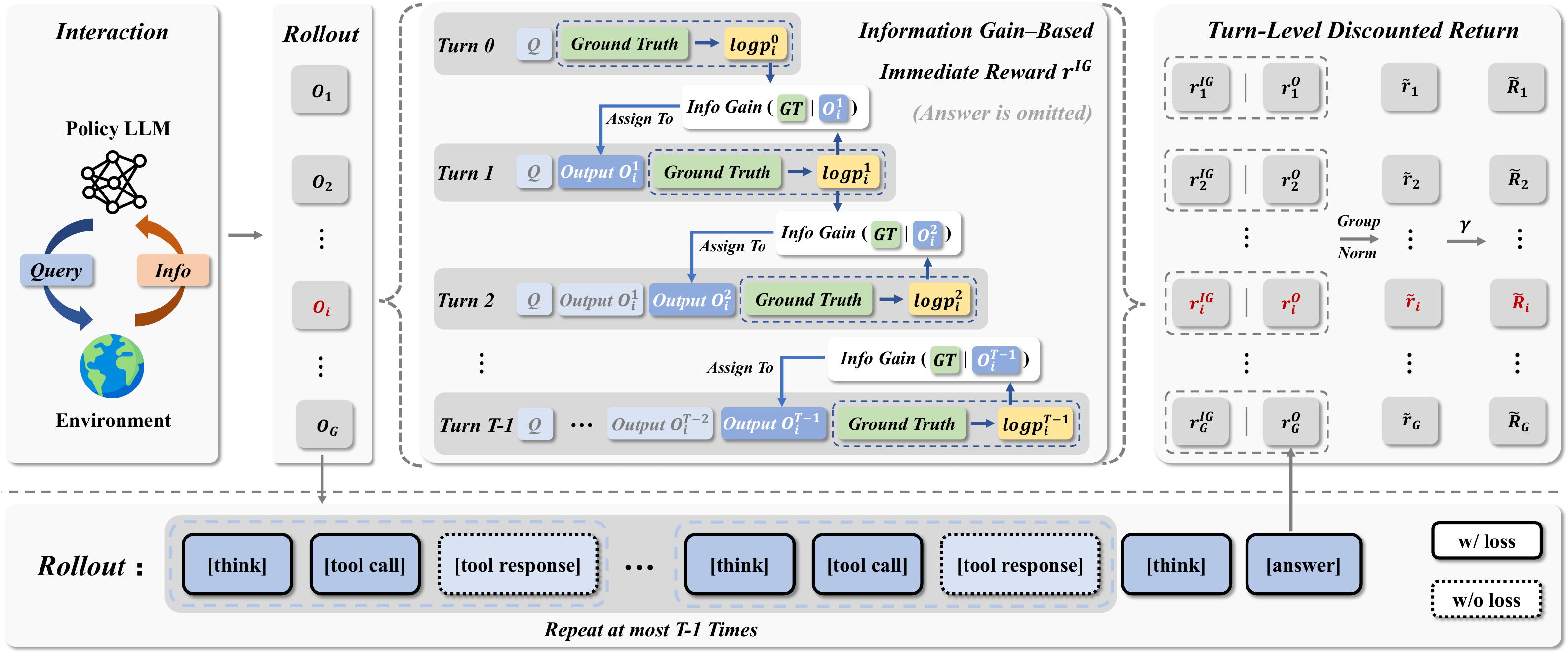}
    \caption{The training pipeline of IGPO. (Upper) Turn-level information gain rewards are computed by measuring changes in ground-truth probability and combined with the outcome reward to derive discounted returns. (Lower) Each rollout contains at most $T-1$ interaction turns, where each turn includes a reasoning step, a tool call, and the returned tool response, followed by a final answer turn. During optimization,  the loss on tool response is masked out.
}
\vspace{-10pt}
    \label{fig:method}
\end{figure}

Let $\mathcal{D}=\{(q,a)\}$ denote a dataset of question–answer pairs, and let $\mathcal{E}$ represent an external tool (e.g., a web search engine). The goal of the agent is to solve question $q$ by generating a rollout 
$o = (\tau_1, \tau_2, \ldots, \tau_T)$ through iterative interaction with the environment via tool $\mathcal{E}$, where $T$ is the total number of interaction turns. The last turn $\tau_T$ is the \textit{answer turn} that outputs a rationale-then-final answer sequence, while all previous turns involve reasoning and tool interaction.
Specifically, for $t<T$, each turn $\tau_t$ is defined as a triple consisting of \texttt{[think]}, \texttt{[tool call]}, and \texttt{[tool response]}. 
The \texttt{[think]} step compels the agent to reason explicitly before acting, and each reasoning process is wrapped in a \texttt{<think></think>} tag following the DeepSeek-R1 setting~\citep{guo2025deepseek}. 
The \texttt{[tool call]} step invokes the external tool $\mathcal{E}$ by producing a structured request, typically JSON-formatted and wrapped in a dedicated tag (e.g., \texttt{<tool\_call>search query</tool\_call>}). 
The \texttt{[tool response]} step then returns structured outputs from $\mathcal{E}$, such as webpage snippets with titles, URLs, and text when using a web search engine tool, enclosed in \texttt{<tool\_response>retrieved documents</tool\_response>} tags. 
In the final turn, after a \texttt{[think]} step, the agent generates its answer within the \texttt{<answer></answer>} tag, and this content is extracted as the trajectory’s final prediction $\hat{a}$, which is expected to correctly address the input query $q$. This agent-environment interaction is illustrated at the bottom of~\autoref{fig:method}.

\subsection{Agentic Reinforcement Learning Pipeline}
\label{Agentic RL Pipline}
\textbf{Policy Optimization.}
Agentic RL typically adopts policy-gradient methods to optimize the agent policy $\pi_\theta$.
A common approach is \emph{Group Relative Policy Optimization} (GRPO)~\citep{shao2024deepseekmath}, which removes the need for an explicit critic by normalizing returns within each sampled group of rollouts. 
Formally, given an actor model $\pi_\theta$, a group of $G$ rollouts $\{o_i\}_{i=1}^G$ is sampled from old policy $\pi_{\theta_{\text{old}}}(\cdot \mid q)$ for each input $(q,a)\sim\mathcal{D}$. The policy is then optimized by maximizing the clipped surrogate objective with KL regularization:
\begin{equation}
\label{eq:grpo}
\begin{aligned}
\mathcal{J}_{\mathrm{GRPO}}(\theta) 
=\; & 
\mathbb{E}_{(q,a)\sim\mathcal{D},\,\{o_i\}\sim \pi_{\theta_{\text{old}}}(\cdot|q)}
\Bigg[
\frac{1}{G}\sum_{i=1}^G \frac{1}{|o_i|}\sum_{t=1}^{|o_i|}
\min\!\Bigg(
\frac{\pi_\theta(o_{i,t}\mid q,o_{i,<t})}{\pi_{\theta_{\text{old}}}(o_{i,t}\mid q,o_{i,<t})}\,\widehat{A}_i, \\[4pt]
& \qquad\qquad
\mathrm{clip}\!\left(\frac{\pi_\theta(o_{i,t}\mid q,o_{i,<t})}{\pi_{\theta_{\text{old}}}(o_{i,t}\mid q,o_{i,<t})},\,1-\epsilon,\,1+\epsilon\right)\,\widehat{A}_i
\Bigg)
-\beta\,\mathbb{D}_{\mathrm{KL}}(\pi_\theta \,\|\, \pi_{\mathrm{ref}})
\Bigg],
\end{aligned}
\end{equation}
where 
$\widehat{A}_{i} = \tfrac{r_{i} - \mathrm{mean}(r_{1},r_{2},\cdots, r_{G})}{\mathrm{std}(r_{1},r_{2},\cdots, r_{G})}$ 
is the normalized group-relative advantage for the $i$-th rollout and $r_i$ is the outcome reward of the $i$-th rollout. $\epsilon$ is the clipping ratio, and $\beta$ controls the KL penalty relative to the reference model $\pi_{\mathrm{ref}}$. 
During optimization, gradients are applied only to decision tokens (reasoning, tool calls, answers), while tool responses from the environment are masked out.

\textbf{Reward.}
During training, the agent receives a scalar reward $r$ for each rollout $o$, which provides the optimization signal. 
Prior work typically combines an outcome reward with a format penalty:
\begin{equation}
\label{eq:reward}
r^O =
\begin{cases}
\;\;\mathrm{F1}(\hat{a}, a) = \frac{2\,|\hat{a} \cap a\,|}{|\hat{a}| + |a|}\;\in[0,1], & \text{if the output is in valid format}, \\[6pt]
\;\;\lambda_{\text{fmt}}, & \text{otherwise},
\end{cases}
\end{equation}
where $\hat{a}$ is the predicted answer, $a$ is the ground-truth answer, and $\mathrm{F1}(\hat{a}, a)\in[0,1]$ denotes the word-level F1 score. 
If the output violates the required schema (e.g., missing tags or malformed JSON), a negative constant $\lambda_{\text{fmt}}<0$ is assigned as a penalty. Thus, the outcome reward provides a correctness signal, while the format penalty enforces the structural validity of outputs.

\section{Information Gain-based Policy Optimization}
In this section, we first illustrate our motivation and then provide a detailed introduction to our proposed information gain-based policy optimization, whose overall framework is shown in~\autoref{fig:method}.

\subsection{Motivation}

While outcome-based RL is effective in single-turn tasks, extending it to multi-turn agentic settings faces three critical limitations. First, standard GRPO leads to \emph{advantage collapse}. In the standard framework (Eq.~\ref{eq:grpo}), each rollout $o_i$ receives a scalar reward derived solely from the final answer. For complex (or trivial) queries, rollouts often yield identical outcomes (uniformly zero or one), causing group-relative advantages to vanish and providing no valid gradient signal. Second, outcome-only supervision \emph{lacks fine-grained credit assignment}. In multi-turn scenarios, later decisions strictly depend on earlier ones: a tool call may be conceptually correct but rendered useless by prior errors, or conversely, valid reasoning may be overshadowed by subsequent mistakes. Outcome rewards obscure these dependencies, failing to distinguish productive steps from invalid ones. Third, outcome reward sparsity results in \emph{poor sample efficiency}. By relying solely on a single terminal signal, the dense semantic information embedded in intermediate reasoning and tool interactions is wasted, necessitating significantly more samples to learn effective policies

To mitigate this issue, we introduce Information-Gain-based Policy Optimization (IGPO). 
The key idea is to exploit the multi-turn structure of agentic rollouts and treat each turn as an opportunity to acquire additional evidence toward the ground truth. At every turn, IGPO measures the increase in the policy’s confidence of generating the correct answer, which we defined as the \emph{information gain} of this turn and uses this as the turn-level reward. By rewarding turn-level information gain, IGPO supplies denser and more fine-grained supervision, especially at early training stages. We further present a theoretical analysis in~\autoref{appendix:theory}, which intuitively explains why IGPO effectively addresses the limitations of sparse outcome rewards in multi-turn scenarios.
Since the information gain is defined with respect to the ground-truth answer and computed under teacher forcing, it provides rich and dense supervision, ensuring that every sample contributes to learning even when no rollout produces a fully correct final answer.

\subsection{Information Gain-based Turn-level Reward}
\label{Sec-info gain reward}
\textbf{Information Gain Reward.}
We view multi-turn agent–environment interaction as a process of \emph{incrementally acquiring information about the ground truth}. 
To capture this intuition, we propose an intrinsic \emph{information gain-based reward}. 
At each turn, we evaluate the policy’s probability of generating the ground-truth answer and define the reward as the difference between consecutive states. 
We call this the \emph{information gain reward}, as it measures the \emph{marginal increase in posterior probability mass assigned to the ground truth} induced by the current turn. In practice, to ensure numerical stability, we quantify this gain as the increment in log probability.

Formally, let $a=(a_1,\ldots,a_L)$ denote the ground-truth answer tokens. 
For the $t$-th turn in the $i$-th rollout, the log probability of $a$ under the current policy $\pi_\theta$ is computed as
\begin{equation}
\label{eq:logp}
\log \pi_\theta(a \mid q, o_{i,\leq t})
=
\frac{1}{L}\sum_{j=1}^{L}
\log \pi_\theta(a_j\mid q, o_{i,\leq t},a_{<j})
,
\end{equation}
where $o_{i,\leq t}$ denotes the prefix of rollout $o_i$ up to turn $t$. 
Then the immediate reward~\footnote{Due to its log-prob origin, we apply stop-gradient to the information gain–based reward.} for turn $t$ is
\begin{equation}
\label{eq:info gain reward}
r_{i,t}^{IG}=\mathrm{IG}(a\mid q, o_{i,t})
=\log \pi_\theta(a \mid q, o_{i,\leq t})-\log \pi_\theta(a \mid q, o_{i,\leq t-1}),\qquad 1\leq t<T.
\end{equation}
In practice, the ground-truth answer $a$ is wrapped in the same schema as a predicted answer to ensure consistency with rollout formatting, e.g.,  
\texttt{<think>Now there's enough information to answer</think><answer>Ground Truth $a$</answer>}.

This turn-level reward offers three key properties: (1) \emph{Ground-truth awareness}: It increases when the turn-action raises the policy’s confidence in the correct answer, and decreases otherwise. Crucially, this objective derivation minimizes the potential bias inherent in external reward models or manual process labeling. (2) \emph{Dense supervision}: It provides turn-level dense signals that resolve advantage collapse, enable fine-grained credit assignment, and improve sample efficiency. (3) \emph{Computational efficiency}: Unlike other process reward designs (especially Monte Carlo estimation), it incurs negligible overhead, an advantage further amplified by the vectorized implementation presented below

\begin{wrapfigure}{r}{0.44\textwidth} 
    \vspace{-5pt}
    \centering
    \includegraphics[width=0.44\textwidth]{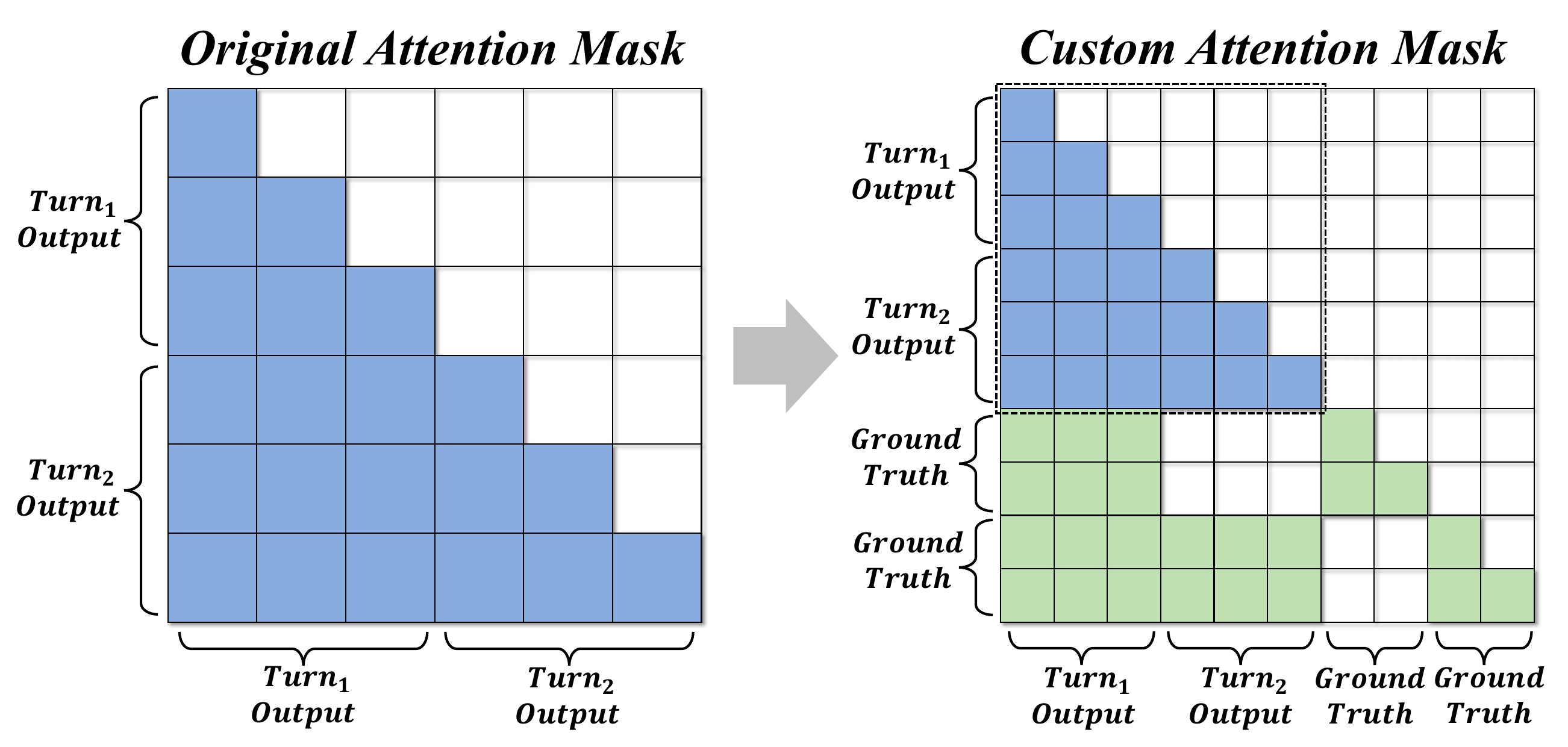}
    \vspace{-10pt}
    \caption{The custom attenyion mask for vectorized implementation. Shaded cells denote allowed attention. Prompt and answer tokens are omitted for clarity.}
    \vspace{-10pt}
    \label{fig:mask}
\end{wrapfigure}
\textbf{Efficient Vectorized Implementation.} 
Although the information gain reward already enjoys inherent efficiency over existing process reward methods, we seek to further optimize its computation. In the standard implementation, computing the information gain reward requires $T$ separate forward passes, with complexity scaling as $\sum_{t=0}^{T-1} L_t^2$, where $L_t$ is the sequence length at turn $t$. To optimize this, we propose a vectorized implementation that appends $T$ copies of the ground-truth answer to the end of the trajectory. By applying a custom attention mask (\autoref{fig:mask}) that restricts each copy's visibility to its corresponding turn, we compute the log probabilities for all T turns simultaneously in a single forward pass (complexity $\propto L_{T-1}^2$), while ensuring mathematical equivalence to Eq.~\ref{eq:logp}. Given that the ground-truth length is orders of magnitude smaller than the full reasoning trajectory, the overhead introduced by these appended copies is negligible. This implementation yields an asymptotic speedup of $\approx T/3$ (e.g., $3\times$ for $T=10$) and further enhances GPU utilization by reducing synchronization overhead.

\subsection{Policy Optimization with Turn-Level Discounted Return}

\textbf{Turn-Level Discounted Return.}

Given a group of rollouts $\{o_i\}_{i=1}^G$, where each rollout $o_i$ yields a sequence of $T-1$ information gain rewards $\{r_{i,t}^{\text{IG}}\}_{t=1}^{T-1}$ and an outcome reward $r_i^{\text{O}}$ (calculated via Eq.~\ref{eq:reward}), we follow GRPO~\citep{shao2024deepseekmath} to stabilize training and capture the relative magnitude of rewards. Specifically, we perform group-wise $z$-normalization on the information gain rewards and outcome rewards separately.

\begin{equation}
\label{eq:norm}
\tilde{r}_{i,t} =
\begin{cases}
\frac{r_{i,t}^{\text{IG}} - \mu_{\text{IG}}}{\sigma_{\text{IG}}}, & 1 \leq t < T, \\[8pt]
\frac{r_i^{\text{O}} - \mu_{\text{O}}}{\sigma_{\text{O}}}, & t = T,
\end{cases}
\end{equation}
where $\mu_{\text{IG}}, \sigma_{\text{IG}}$ are the mean and standard deviation of all information gain rewards $\{r_{i,t}^{\text{IG}}\}$ within the group, and $\mu_{\text{O}}, \sigma_{\text{O}}$ correspond to the outcome rewards $\{r_i^{\text{O}}\}$.

While $\tilde{r}_{i,t}$ captures the relative quality of each turn, it only reflects immediate effects and ignores the impact of current decisions on future turns.  To incorporate such long-horizon dependencies, we compute a turn-level discounted return $\tilde{R}_{i,t}$ to reflect the cumulative impact of future rewards:
\begin{equation}
\label{eq:disc-adv}
\tilde{R}_{i,t} = \sum_{k=t}^{T} \gamma^{\,k-t} \tilde{r}_{i,k},
\end{equation}
where $\gamma \in (0,1]$ is the discount factor. 
During optimization, $\tilde{R}_{i,t}$ is assigned to all tokens produced in turn $t$ of rollout $o_i$. This yields a dense and future-aware supervision signal for policy learning.

\textbf{Policy Optimization.}
With the turn-level discounted return $\tilde{R}_{i,t}$ defined above, we optimize the agent policy using a clipped surrogate objective with KL regularization, following the same structure as GRPO but with a finer-grained credit assignment. 
Formally, the IGPO objective is
\begin{equation}
\label{eq:igpo}
\begin{aligned}
\mathcal{J}_{\mathrm{IGPO}}(\theta) 
=\; & 
\mathbb{E}_{(q,a)\sim\mathcal{D},\,\{o_i\}\sim \pi_{\theta_{\text{old}}}(\cdot\mid q)}
\Bigg[
\frac{1}{G}\sum_{i=1}^{G}\frac{1}{|o_i|}
\sum_{t=1}^{|o_i|}
\min\!\Bigg(
\frac{\pi_\theta(o_{i,t}\mid q,o_{i,<t})}{\pi_{\theta_{\text{old}}}(o_{i,t}\mid q,o_{i,<t})}\, \color{black} \tilde{R}_{i,t}, \\[4pt]
& \qquad\qquad
\mathrm{clip}\!\left(
\frac{\pi_\theta(o_{i,t}\mid q,o_{i,<t})}{\pi_{\theta_{\text{old}}}(o_{i,t}\mid q,o_{i,<t})},\,1-\epsilon,\,1+\epsilon
\right)\, \color{black} \tilde{R}_{i,t} \color{black}
\Bigg)
-\beta\,\mathbb{D}_{\mathrm{KL}}(\pi_\theta \,\|\, \pi_{\mathrm{ref}})
\Bigg],
\end{aligned}
\end{equation}
where $\epsilon$ is the clipping threshold, $\beta$ controls the KL penalty strength, and $t$ maps token $o_{i,t}$ to its originating turn. 
During optimization, only decision tokens (reasoning, tool calls, and answers) receive gradient updates, while raw tool responses are masked out.

To further substantiate the simplicity and implementability of the proposed IGPO, we provide an algorithmic flow comparison between IGPO and GRPO in \autoref{appx:grpo vs igpo}.

\section{Experiments}

\subsection{Experimental Setup}

\textbf{Datasets \& Metrics.} To evaluate the effectiveness of our proposed IGPO, we conduct experiments on both in-domain (ID) and out-of-domain (OOD) QA benchmarks in an agentic search setting. Following previous work~\citep{zheng2025deepresearcher, deng2025atom}, the ID setting includes four widely used datasets: NQ~\citep{kwiatkowski2019natural}, TQ~\citep{joshi2017triviaqa}, HotpotQA~\citep{yang2018hotpotqa}, and 2Wiki~\citep{ho2020constructing}, while the OOD setting includes three datasets: MusiQue~\citep{trivedi2022musique}, Bamboogle~\citep{press2022measuring}, and PopQA~\citep{mallen2022not}.
We report word-level F1 as the evaluation metric, which is computed as the harmonic mean of precision and recall between the predicted and reference answers.  

\textbf{Baselines.} To directly verify IGPO’s superiority on agentic search tasks, we compare it against a set of competitive baselines: (1) Prompt-based methods: CoT~\citep{wei2022chain}, CoT+RAG~\citep{gao2023retrieval}, and Search-o1~\citep{li2025search}, which represent the baseline performance of LLMs without further training on search tasks. (2) Outcome-reward RL-based methods: Search-r1-base/Instruct \citep{jin2025search}, R1-searcher \citep{song2025r1}, and DeepResearcher \citep{zheng2025deepresearcher}, the representative search agents with outcome-based reward RL, yielding marked performance gains. (3) Step-reward RL-based methods: StepSearch-base/instruct \citep{wang2025stepsearch}, ReasoningRAG \citep{zhang2025process}, and GiGPO \citep{feng2025group}, which are the latest approaches exploring step-reward RL in search-agent settings.

To further validate IGPO’s effectiveness, we also compare it against the following commonly used RL algorithms under the same configuration: PPO \citep{schulman2017proximal}, a widely used actor-critic algorithm that requires an additional value model, and critic-free methods Reinforce++ \citep{hu2025reinforce++}, RLOO \citep{kool2019buy, ahmadian2024back}, GRPO \citep{shao2024deepseekmath}, and GSPO\citep{zheng2025group} which perform advantage estimation over trajectory groups or batchs.

\textbf{Implementation Details.}
We use Qwen2.5-7B-Instruct \citep{qwen2025qwen25technicalreport} as our backbone model. The training is conducted using the verl~\citep{Sheng_2025} framework. The discounted factor $\gamma$ is set to 1.0 with no future tuning. At each training step, we sample 32 prompts, and sample 16 rollouts for each prompt. The maximum dialogue turns are set to 10. For the environment, we use the google search API as our tool. The settings of our experiments are consistent with DeepResearcher~\citep{zheng2025deepresearcher}. For the other baselines in~\autoref{tab:search_result}, we directly copy their reported results. All RL training methods
(including ours and the baselines) use exactly the same hyperparameter configurations. The training and inference prompt templates are shown in \autoref{app:prompt}. Please refer to \autoref{A: training details} for more details.

\begin{table}[t]
\setlength{\tabcolsep}{1.6mm}
\centering
\caption{Main results of IGPO compared with different agentic RL baselines across seven datasets.}
\label{tab:search_result}
\begin{tabular}{@{}lcccclcccc@{}}
\toprule
                    & \multicolumn{4}{c}{In-domain}                                 &           & \multicolumn{3}{c}{Out-of-domain}             &               \\ \cmidrule(lr){2-5} \cmidrule(lr){7-9}
Method              & NQ            & TQ            & HotpotQA      & 2Wiki         & \textbf{} & Musique       & Bamboogle     & PopQA         & Avg.          \\ \midrule
\multicolumn{10}{l}{\cellcolor[HTML]{EFEFEF}\textbf{Prompt-based}}                                                                                              \\
CoT                 & 19.8          & 45.6          & 24.4          & 26.4          &           & 8.5           & 22.1          & 17.0          & 23.4          \\
CoT+RAG             & 42.0          & 68.9          & 37.1          & 24.4          &           & 10.0          & 25.4          & 46.9          & 36.4          \\
Search-o1           & 32.4          & 58.9          & 33.0          & 30.9          &           & 14.7          & 46.6          & 38.3          & 36.4          \\ \midrule
\multicolumn{10}{l}{\cellcolor[HTML]{EFEFEF}\textbf{Outcome-reward RL-based}}                                                                                   \\
Search-r1-base      & 45.4          & 71.9          & \underline{55.9}          & 44.6          &           & 26.7          & 56.5          & 43.2          & 49.2          \\
Search-r1-instruct  & 33.1          & 44.7          & 45.7          & 43.4          &           & 26.5          & 45.0          & 43.0          & 40.2          \\
R1-searcher         & 35.4          & 73.1          & 44.8          & 59.4          &           & 22.8          & 64.8          & 42.7          & 49.0          \\
DeepResearcher      & 39.6          & \underline{78.4}          & 52.8          & \underline{59.7}          &           & 27.1          & \underline{71.0}          & \underline{48.5}          & \underline{53.9}          \\ \midrule
\multicolumn{10}{l}{\cellcolor[HTML]{EFEFEF}\textbf{Step-reward RL-based}}                                                                                      \\
StepSearch-base     & -             & -             & 49.3          & 45.0          &           & \underline{32.4} & 57.3          & -             & 46.0          \\
StepSearch-instruct & -             & -             & 50.2          & 43.1          &           & 31.2          & 53.4          & -             & 44.5          \\
ReasoningRAG        & -             & -             & 48.9          & 50.4          &           & 20.6          & 45.5          & 46.2          & 42.3          \\
GiGPO               & \textbf{46.4}          & 64.7          & 41.6          & 43.6          &           & 18.9          & 68.9          & 46.1          & 47.2          \\
\textbf{IGPO} & \textbf{46.4} & \textbf{80.6} & \textbf{59.0} & \textbf{72.1} &           & \textbf{32.7}          & \textbf{77.0} & \textbf{53.8} & \textbf{60.2} \\ \bottomrule
\end{tabular}
\vspace{-0.2cm}
\end{table}

\subsection{Overall Performance}
The overall performance comparison between IGPO and the baseline methods is presented in \autoref{tab:search_result} and \autoref{tab:rl_result}.  Based on these results, we can draw the following key observations: 

\textbf{Training-based methods consistently outperform prompt-based baselines.} As shown in \autoref{tab:search_result}, all RL–based methods, whether outcome- or step-reward driven, achieve substantially higher performance than all prompt-based approaches. This confirms that explicit policy optimization is essential for developing effective LLM-based agents, as opposed to relying on prompting alone.

\textbf{Existing step-reward methods yield competitive but unstable improvements compared to outcome-reward RL methods.} While step-reward baselines occasionally surpass outcome-reward ones on specific datasets (e.g., StepSearch on Musique), their overall performance still lags behind the strongest outcome-reward methods such as DeepResearcher. This suggests that existing step-reward designs, although able to provide intermediate guidance, often suffer from noisy or weak supervision signals that limit their generalizability.

\textbf{IGPO achieves the best overall performance across both in-domain and out-of-domain datasets.} Our IGPO outperforms all baselines, with an average score of 60.2, a clear margin over the best method (+6.3 over DeepResearcher). This improvement is attributed to IGPO’s information gain-based reward design, which assigns intrinsic, ground-truth-aware credit at every turn while preserving the outcome reward. By providing fine-grained credit assignment and improves sample efficiency, IGPO delivers robust gains across both in-domain and out-of-domain datasets.

\begin{table}[t]
\centering
\caption{Main results of IGPO compared with different RL baselines across seven datasets.}
\resizebox{1.0 \columnwidth}{!}{%
\begin{tabular}{@{}lccccccccc@{}}
\toprule
              & \multicolumn{4}{c}{In-domain}                                 &  & \multicolumn{3}{c}{Out-of-domain}             & \multicolumn{1}{l}{} \\ \midrule
Method        & NQ            & TQ            & HotpotQA      & 2Wiki         &  & Musique       & Bamboogle     & PopQA         & Avg.                 \\ \midrule
RLOO          & 40.7          & 72.5          & 49.6          & 55.0          &  & 24.8          & 62.2          & 43.1          & 49.7                 \\
PPO           & 38.7          & 75.4          & 48.6          & 59.7          &  & {\ul 26.2}    & 63.4          & {\ul 48.7}    & 51.5                 \\
GRPO          & 40.3          & 77.0          & 48.9          & 57.7          &  & 25.0          & 65.1          & 49.6          & 51.9                 \\
Reinforce++   & 34.3          & 67.5          & 45.9          & 54.5          &  & 23.7          & 61.2          & 44.3          & 47.3                 \\
GSPO          & {\ul 41.5}    & {\ul 77.7}    & {\ul 46.3}    & {\ul 60.1}    &  & 25.4          & {\ul 67.6}    & 45.4          & {\ul 52.0}           \\
\textbf{IGPO} & \textbf{46.4} & \textbf{80.6} & \textbf{59.0} & \textbf{72.1} &  & \textbf{32.7} & \textbf{77.0} & \textbf{53.8} & \textbf{60.2}        \\ \bottomrule
\end{tabular}%
}
\label{tab:rl_result}
\vspace{-0.4cm}
\end{table}

\textbf{IGPO consistently outperforms other RL algorithms.} Beyond task-specific baselines, \autoref{tab:rl_result} shows that IGPO also achieves the highest overall score among standard RL methods, surpassing RLOO, PPO, Reinforce++, and GSPO. Unlike these methods, which rely solely on sparse outcome rewards, IGPO incorporates turn-level discounted returns to provide denser and more stable supervision, leading to stronger generalization and more efficient training.

\subsection{Ablation Study}
We further conduct ablation experiments to assess the contribution of different reward components. As shown in \autoref{tab:ablation}, we observe:

\textbf{First, using only information gain (IG) turn-level reward or only outcome reward (F1) yields clearly inferior results compared to the full combination.} This highlights the complementary roles of turn-level and outcome-level supervision: the outcome reward enforces alignment with the final task objective but suffers from severe sparsity, whereas the information gain reward offers dense and stable guidance for intermediate steps.

\textbf{Second, IGPO with only IG achieves performance comparable to or even exceeding that of standard GRPO (i.e., IGPO w/ F1).} This demonstrates that IGPO’s information gain reward is not subject to reward hacking. Usually, without outcome supervision, unstable reward designs would quickly collapse. In contrast, our IGPO remains robust because its turn-level signals are intrinsically defined and grounded in the ground truth.
\begin{table}[htbp]
\centering
\caption{Ablation results of IGPO on Qwen2.5-3B/7B-Instruct with different reward designs. IGPO (w/ F1) corresponds to using only outcome rewards, reducing to standard GRPO.}
\label{tab:ablation}
\resizebox{\columnwidth}{!}{%
\begin{tabular}{@{}lccccccccc@{}}
\toprule
\multicolumn{1}{c}{} & \multicolumn{4}{c}{In-domain}                                 & \multicolumn{1}{l}{} & \multicolumn{3}{c}{Out-of-domain}             & \multicolumn{1}{l}{}     \\ \cmidrule(lr){2-5} \cmidrule(lr){7-9}
Method               & NQ            & TQ            & HotpotQA      & 2Wiki         & \multicolumn{1}{l}{} & Musique       & Bamboogle     & PopQA         & \multicolumn{1}{l}{Avg.} \\ \midrule
\multicolumn{10}{l}{\cellcolor[HTML]{EFEFEF}\textbf{Qwen2.5-3B-Instruct}}                                                                                                              \\
IGPO (w/ F1)                & {\ul 31.0}    & {\ul 55.6}    & 27.5          & 29.4          &                      & 12.1          & 35.7          & {\ul 34.9}    & 32.3                     \\
IGPO (w/ IG)          & 29.6          & 54.1          & {\ul 28.1}    & {\ul 37.5}    &                      & {\ul 17.6}    & {\ul 43.8}    & 31.7          & {\ul 34.6}               \\
IGPO (w/ F1+IG)        & \textbf{41.9} & \textbf{69.2} & \textbf{47.8} & \textbf{51.4} & \textbf{}            & \textbf{24.8} & \textbf{58.4} & \textbf{49.0} & \textbf{48.9}            \\ \midrule
\multicolumn{10}{l}{\cellcolor[HTML]{EFEFEF}\textbf{Qwen2.5-7B-Instruct}}                                                                                                              \\
IGPO (w/ F1)                & {\ul 40.3}    & {\ul 77.0}    & 48.9          & 57.7          &                      & 25.0          & 65.1          & {\ul 49.6}    & 51.9                     \\
IGPO (w/ IG)          & 37.3          & 75.2          & {\ul 52.1}    & {\ul 63.3}    &                      & {\ul 28.9}    & {\ul 69.8}    & 47.8          & {\ul 53.5}               \\
IGPO (w/ F1+IG)        & \textbf{46.4} & \textbf{80.6} & \textbf{59.0} & \textbf{72.1} & \textbf{}           & \textbf{32.7} & \textbf{77.0} & \textbf{53.8} & \textbf{60.2}            \\ \bottomrule
\end{tabular}%
}
\vspace{-0.5cm}
\end{table}

\begin{figure}[htbp]
    \centering
    \begin{subfigure}{0.24\textwidth}
        \includegraphics[width=\linewidth]{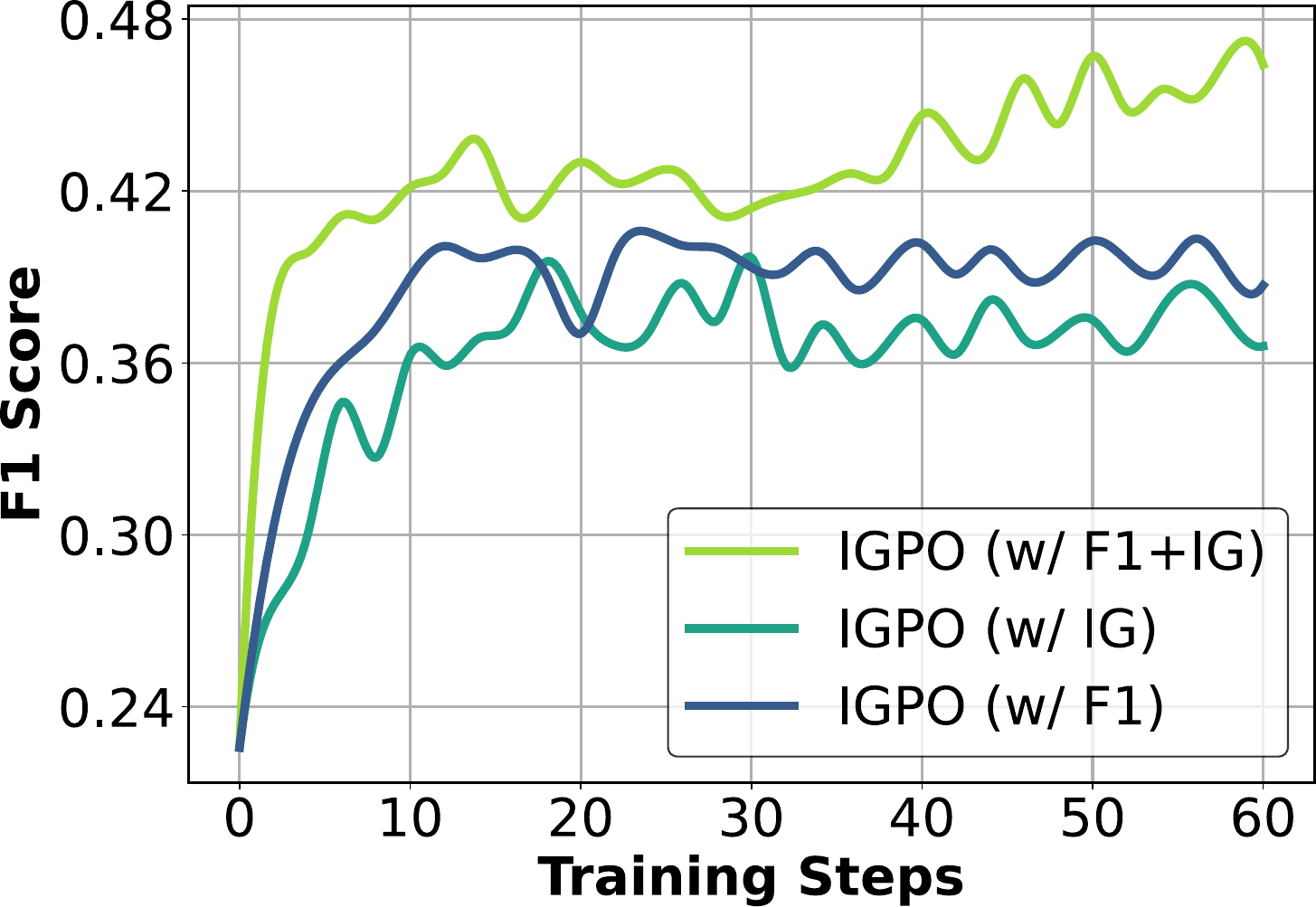}
        \caption{NQ}
    \end{subfigure}
    \begin{subfigure}{0.24\textwidth}
        \includegraphics[width=\linewidth]{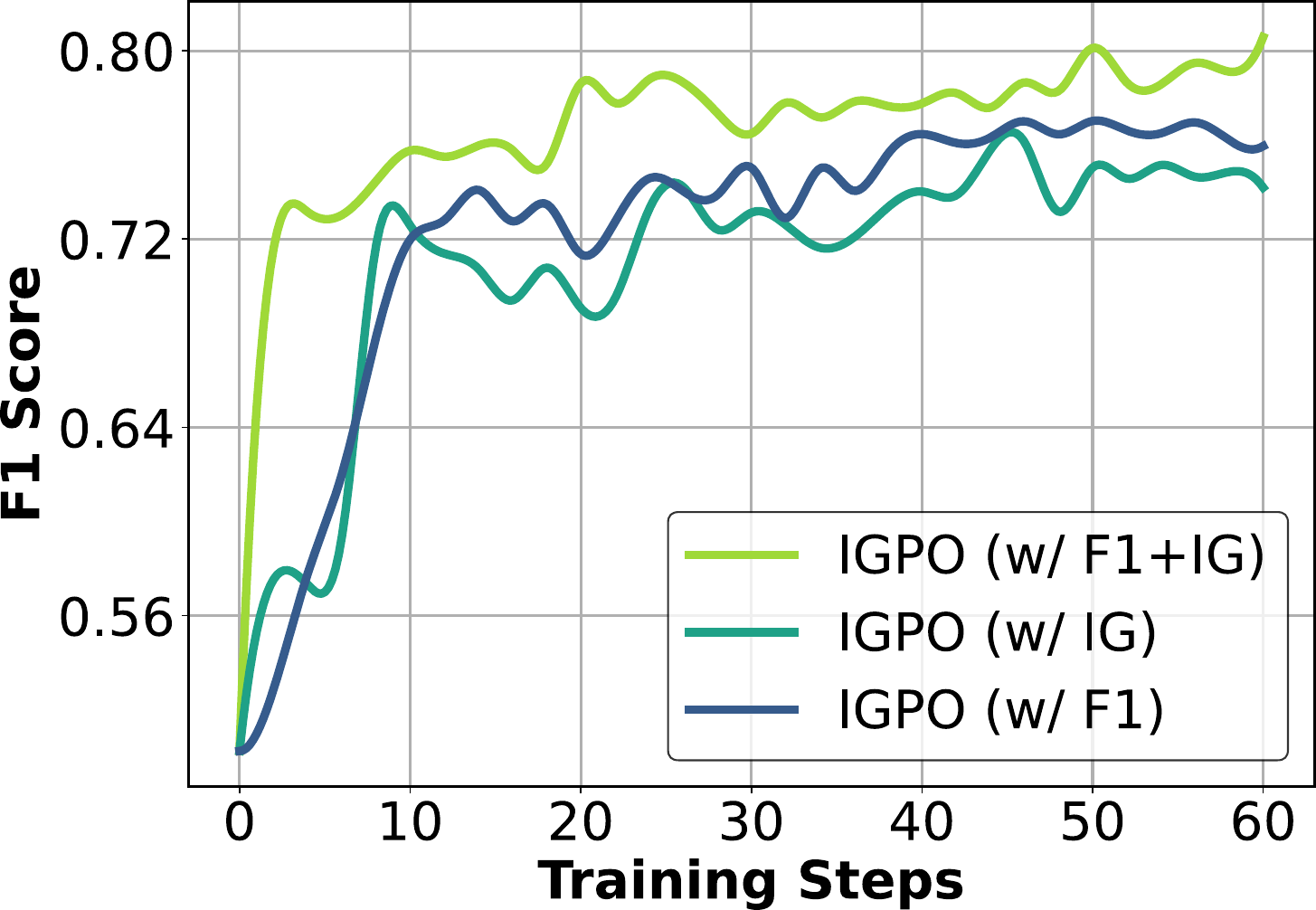}
        \caption{TQ}
    \end{subfigure}
    \begin{subfigure}{0.24\textwidth}
        \includegraphics[width=\linewidth]{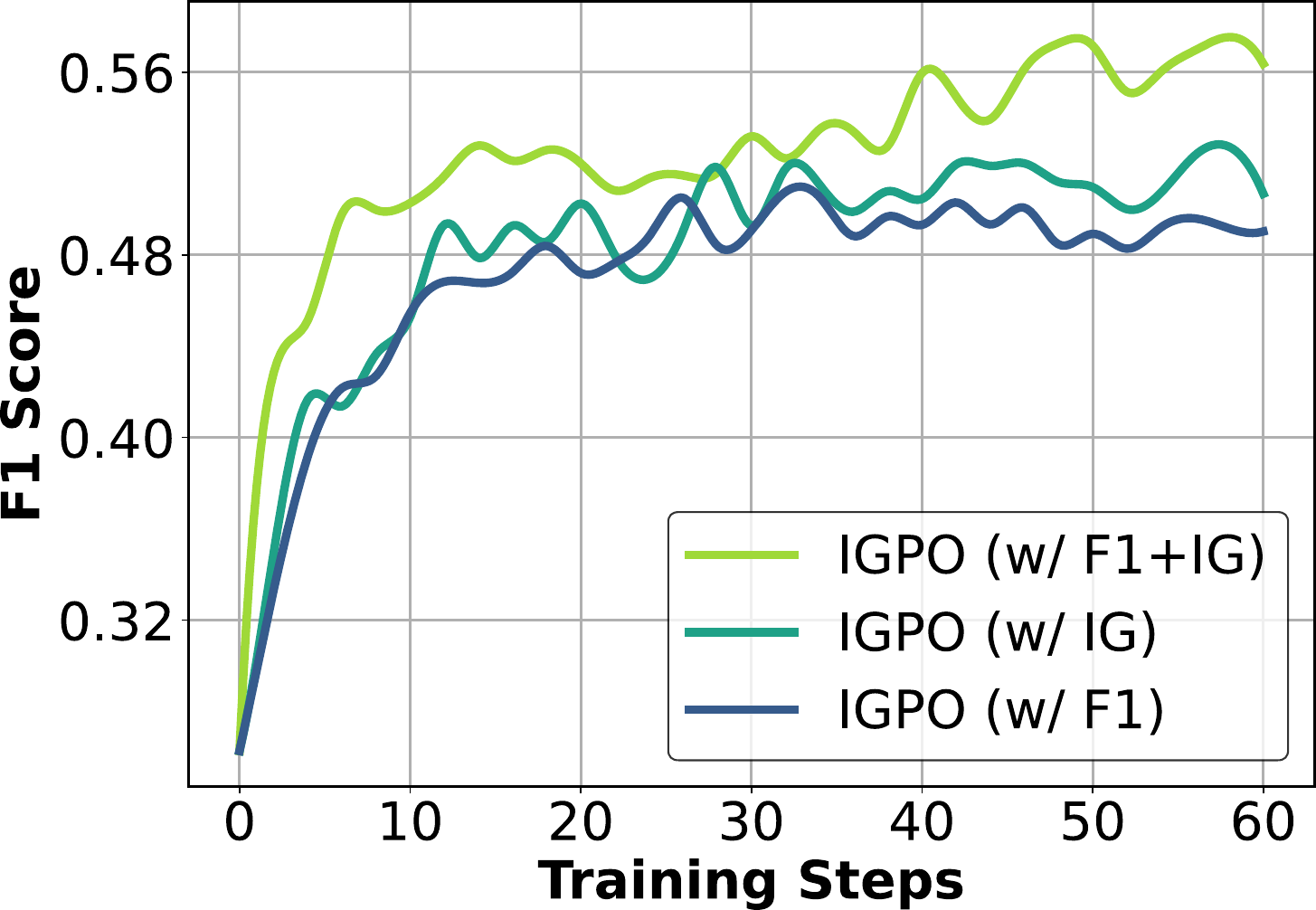}
        \caption{HotpotQA}
    \end{subfigure}
    \begin{subfigure}{0.24\textwidth}
        \includegraphics[width=\linewidth]{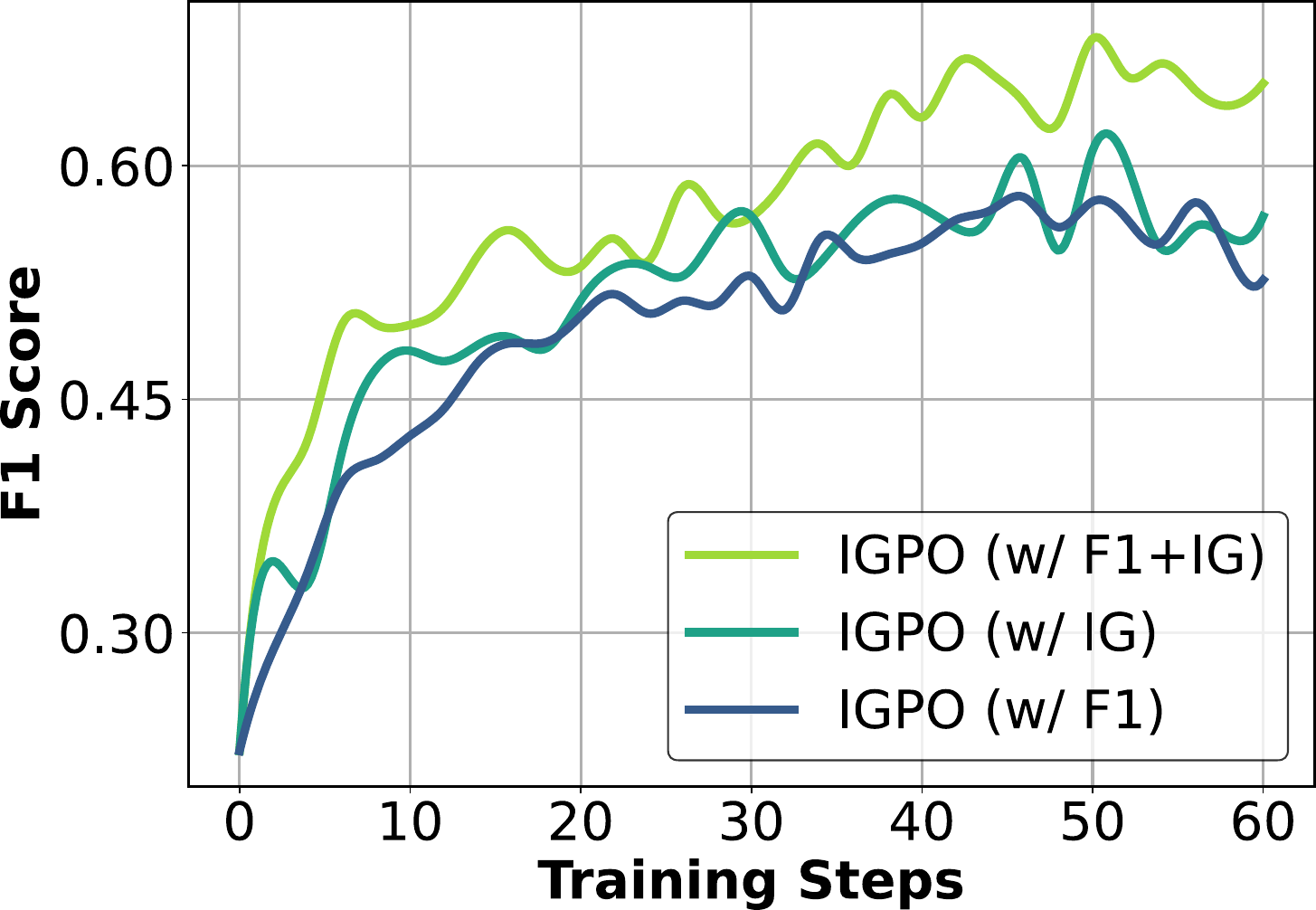}
        \caption{2Wiki}
    \end{subfigure}
    \begin{subfigure}{0.24\textwidth}
        \includegraphics[width=\linewidth]{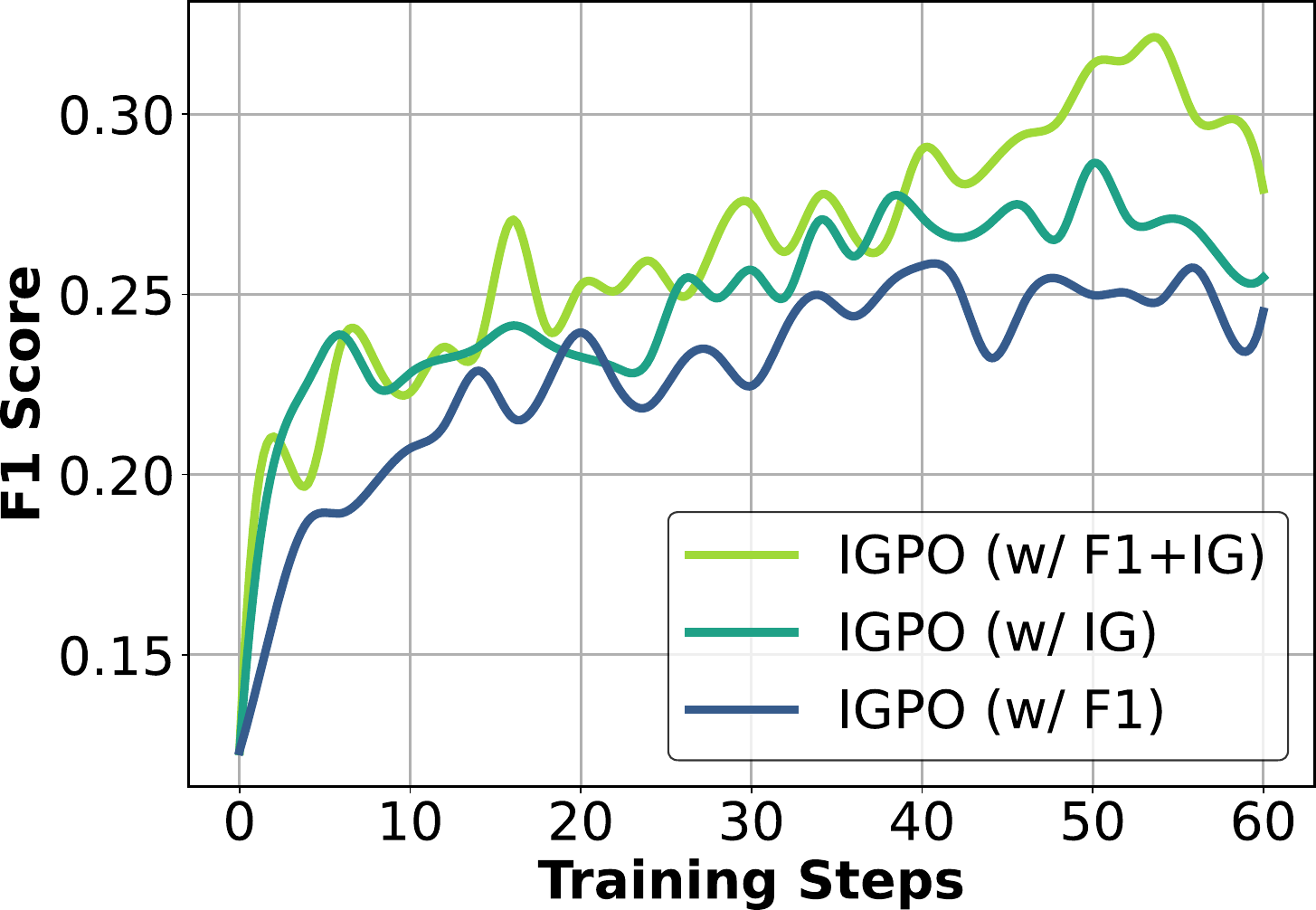}
        \caption{Musique}
    \end{subfigure}
    \begin{subfigure}{0.24\textwidth}
        \includegraphics[width=\linewidth]{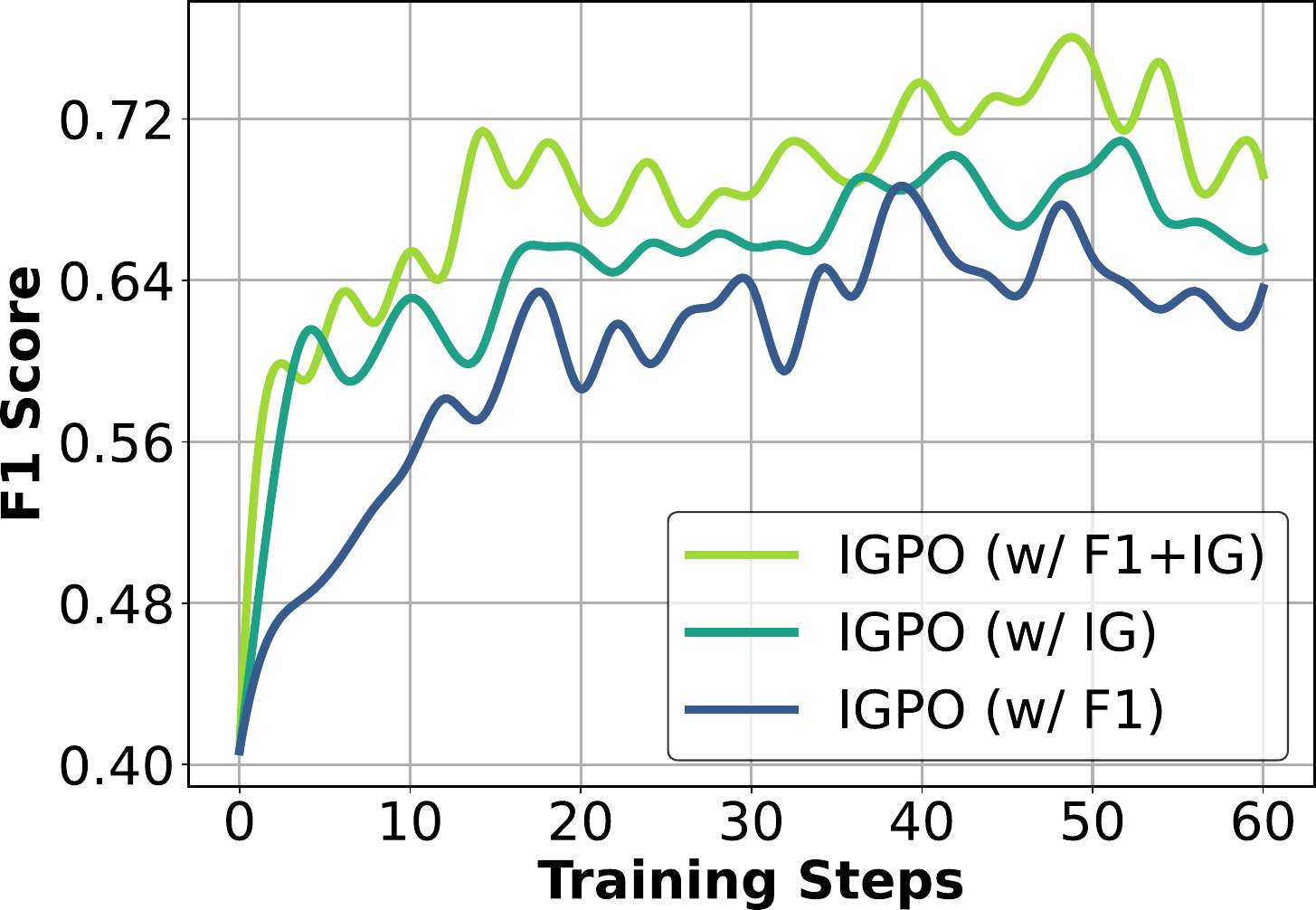}
        \caption{Bamboogle}
    \end{subfigure}
    \begin{subfigure}{0.24\textwidth}
        \includegraphics[width=\linewidth]{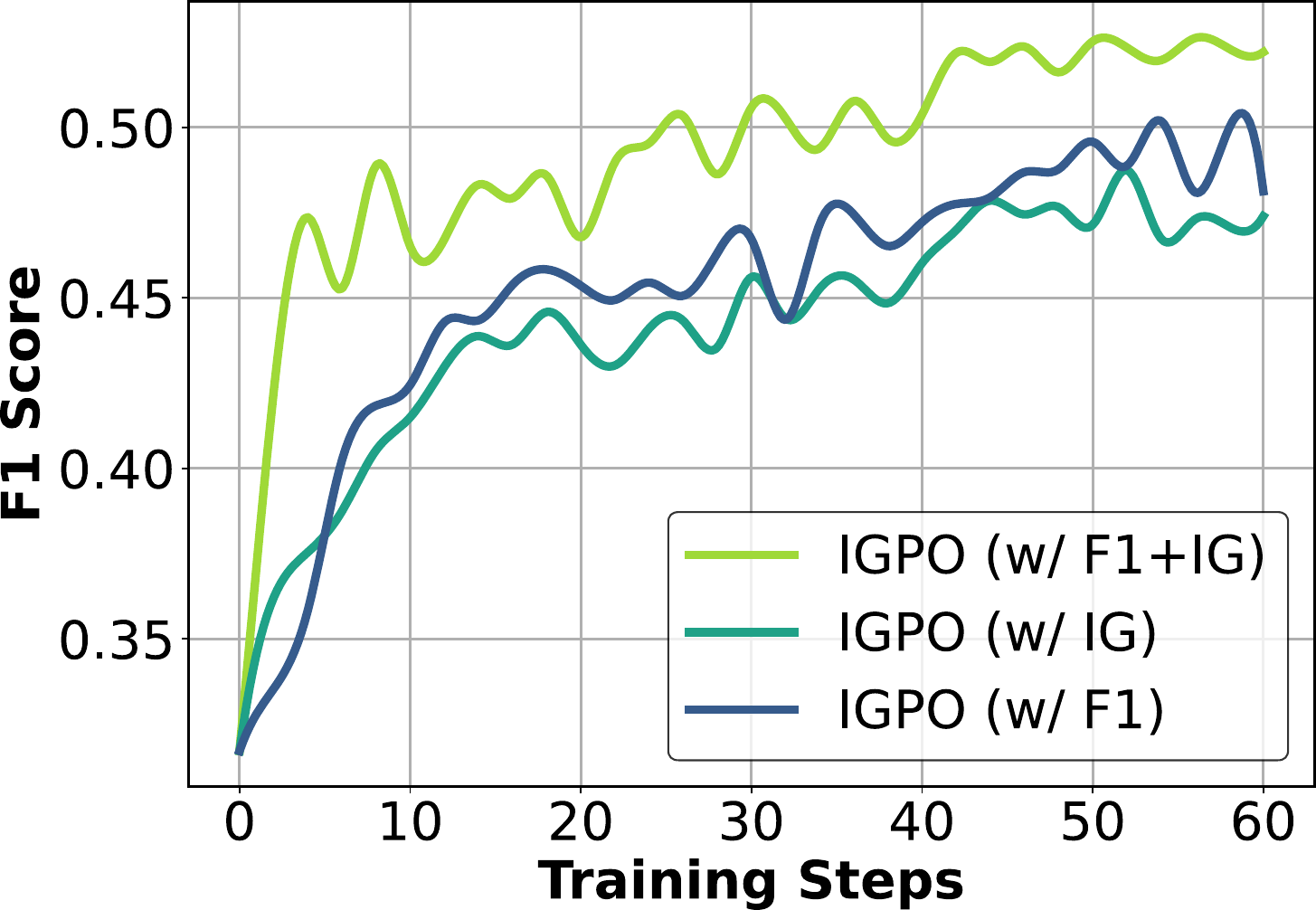}
        \caption{PopQA}
    \end{subfigure}
    \caption{Training curves on Qwen2.5-7B-Instruct with different reward designs.}
    \label{fig:learning_curves}
    \vspace{-0.5cm}
\end{figure}

\textbf{Third, the improvements are particularly pronounced on the smaller 3B model.} Compared to standard GRPO, IGPO improves the 3B model by +16.6 points (32.3 $\rightarrow$ 48.9) and the 7B model by +8.3 points (51.9 $\rightarrow$ 60.2). This larger benefit on the 3B model arises because advantage collapse is more severe for weaker models that struggle to directly produce correct answers (\autoref{fig:adv_collapse}), making them especially reliant on dense reward signals. In such cases, the information gain reward helps prune noisy reasoning paths and reinforce rollouts that progressively approach the ground truth.

\textbf{Finally, IGPO demonstrates consistently faster and more stable learning dynamics.} As shown in ~\autoref{fig:learning_curves}, IGPO steadily outperforms its two ablated variants throughout training across all seven datasets. The curves highlight two advantages: (i) IGPO converges to higher F1 scores, confirming the benefit of combining intrinsic turn-level reward and outcome rewards, and (ii) IGPO maintains stable improvements over steps, indicating robustness against reward sparsity and noisy supervision. These results further validate that IGPO provides dense and reliable training signals, thereby improving both training efficiency and final performance.

In addition to the reward ablation, we compare different information gain bases (probability vs. log-probability) and normalization strategies (joint vs. separate) in Appendix~\ref{info_gain&norm}.

\subsection{In-Depth Analysis}

\begin{figure}[htbp]
    \centering
    \begin{minipage}{0.48\linewidth}
        \centering
        \includegraphics[width=0.95\linewidth]{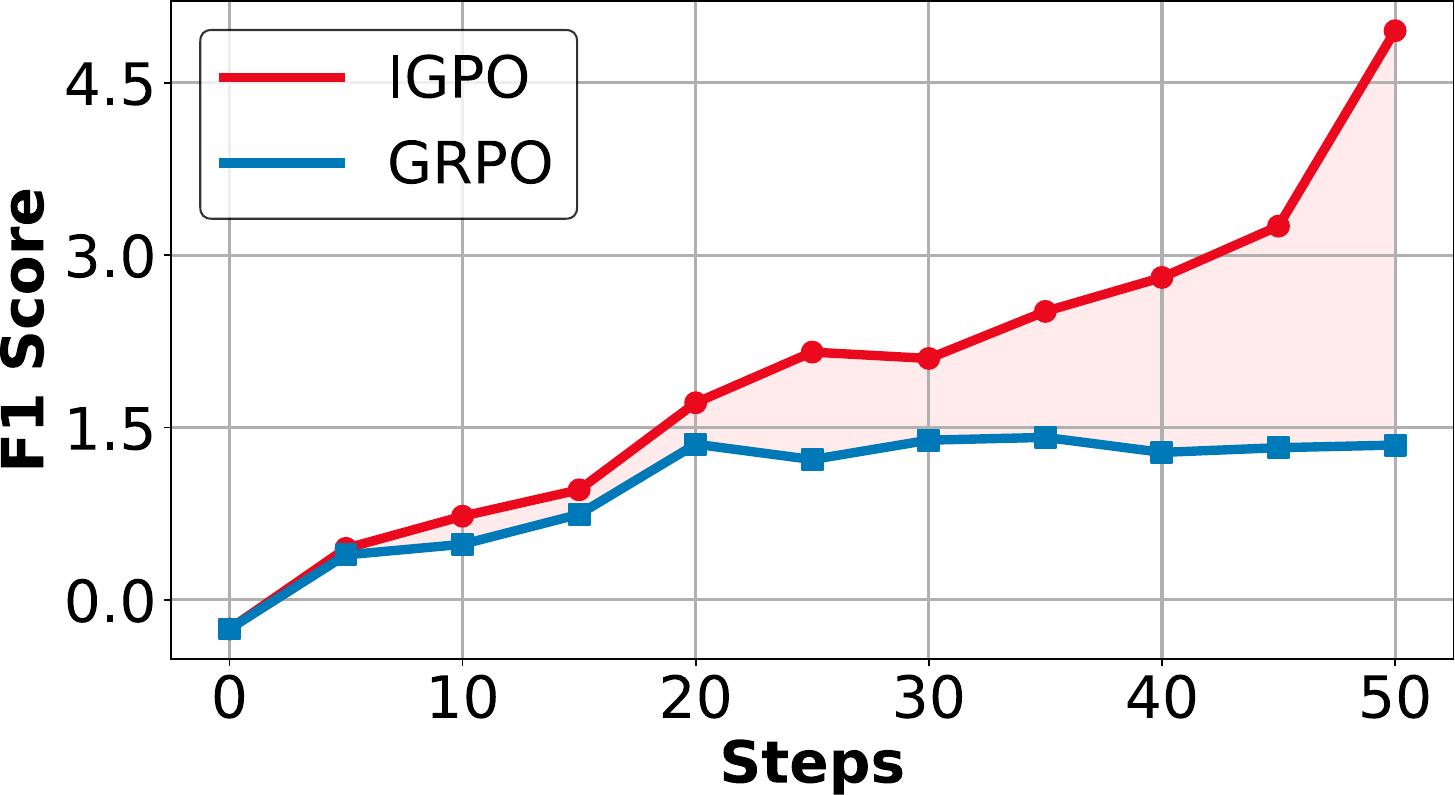}
        \caption{Mean reduction in ground-truth answer entropy from the initial query (Turn 0) to the last non-answer turn ($T\!-\!1$) during training.}
        \label{fig:entropy_reduction}
    \end{minipage}%
    \hfill%
    \begin{minipage}{0.48\linewidth}
        \centering
        \includegraphics[width=0.95\linewidth]{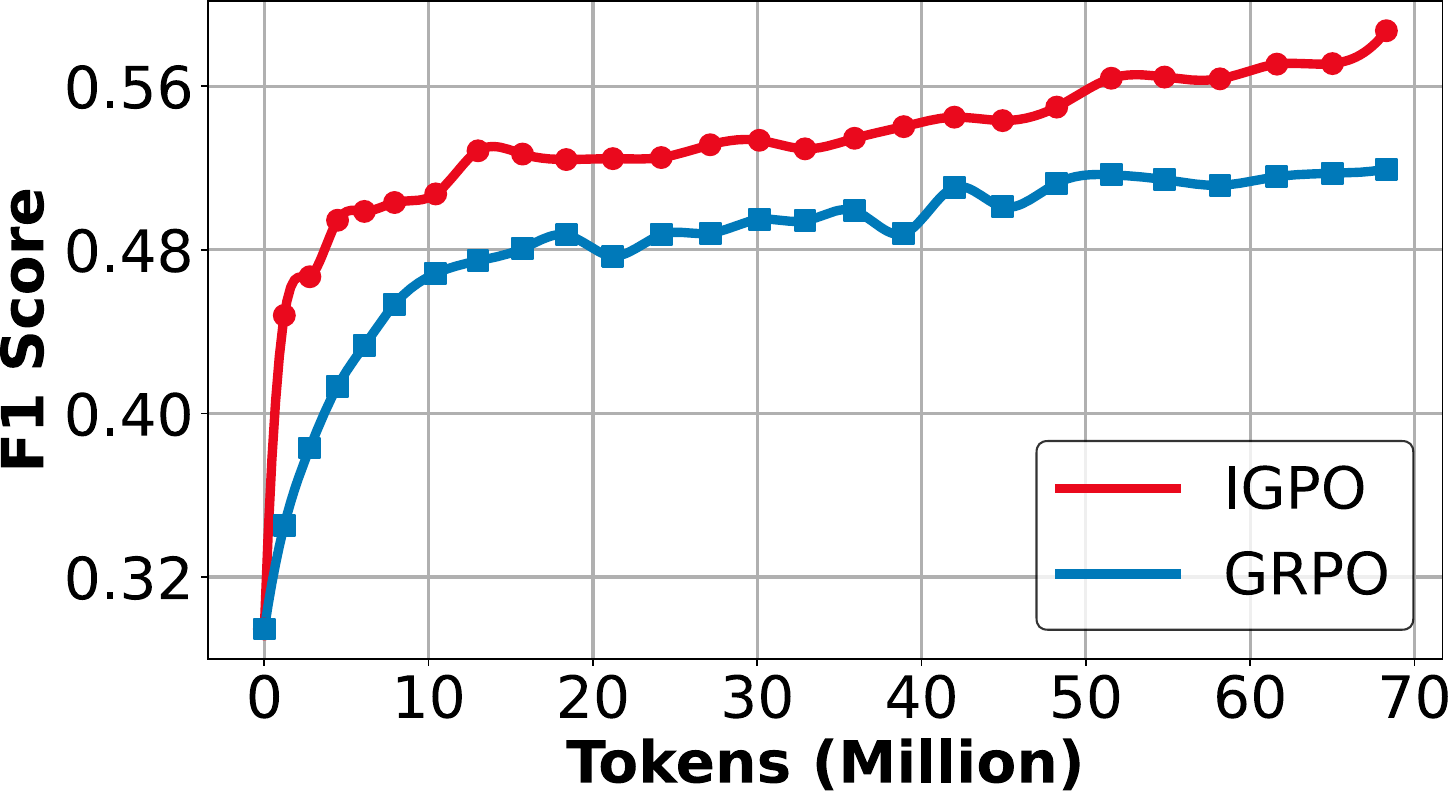}
        \caption{Token Efficiency: average performance with respect to the number of tokens used for gradient updates.}
        \label{fig:token_efficiency}
    \end{minipage}
    \vspace{-0.5cm}
\end{figure}

\paragraph{Ground-truth Entropy Reduction.}
To better understand how IGPO improves training dynamics, we measure the change in ground-truth answer entropy from the initial query (Turn 0) to the last non-answer turn ($T-1$). As shown in \autoref{fig:entropy_reduction}, IGPO consistently achieves a larger entropy reduction than GRPO throughout training. This indicates that the information gain reward effectively encourages intermediate steps to move the policy closer to the ground-truth answer distribution. In contrast, outcome-based supervision in GRPO provides no guidance for intermediate turns, resulting in weaker entropy reduction. These results highlight that IGPO’s turn-level supervision translates into more confident and grounded reasoning trajectories.

\paragraph{Token Efficiency.}
We further compare IGPO and GRPO in terms of token efficiency, i.e., the performance improvement per token used for gradient updates. As shown in \autoref{fig:token_efficiency}, performance increases more rapidly under IGPO, and the gap over GRPO widens as training progresses. In other words, IGPO achieves stronger performance with fewer tokens, indicating that its turn-level rewards deliver denser and more informative gradients than outcome-only supervision, thereby effectively addressing the issue of poor sample efficiency. This finding is consistent with the training dynamics observed in \autoref{fig:learning_curves}, where IGPO not only converges faster but also maintains a stable advantage throughout optimization. Such improvements in token efficiency are particularly valuable in agentic RL, where training data is scarce and expensive to obtain, making efficient use of every gradient update a critical factor for scaling.

The additional experimental analysis is provided in \autoref{app:case study and more discuss}, in particular, the analysis regarding spurious correlations is presented in Appendix~\ref{spurious correlations}. Details regarding IGPO's failure modes are provided in \autoref{failure}. Beyond empirical effectiveness, our theoretical analysis in \autoref{appendix:theory} shows that maximizing turn-level information gain constrains error accumulation in multi-turn scenarios. Thus, IGPO not only alleviates credit-assignment and advantage collapse problems but also reduces error accumulation in long-horizon agentic tasks. The case study can be found in \autoref{case study}.
\vspace{30pt}
\begin{wrapfigure}{r}{0.45\textwidth} 
    \centering
    \includegraphics[width=0.45\textwidth]{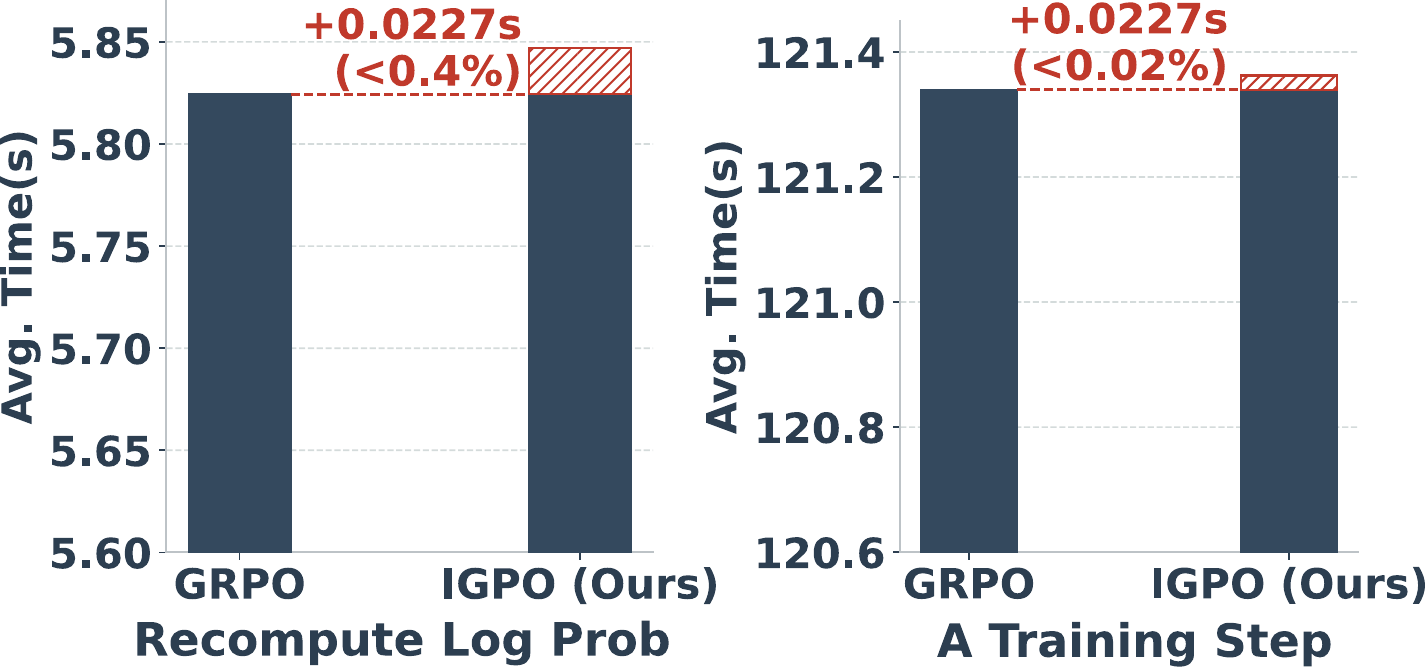}
    \caption{IGPO incurs negligible overhead (0.0227s / step), representing a  \textless 0.4\% increase in info-gain reward computation and \textless 0.02\% end-to-end.}
    
    \label{fig:cost}
\end{wrapfigure}
\vspace{-35pt}
\subsection{Computational Budget}
\label{cost}
We present an empirical runtime comparison between IGPO and GRPO. Results (\autoref{fig:cost}) show that the info-gain reward computation in IGPO incurs an overhead of less than 0.4\% relative to the standard implementation. For the entire training step, IGPO increases latency by less than 0.02\%, demonstrating a training speed nearly identical to GRPO. This underscores the superiority of IGPO: it enhances performance via fine-grained credit assignment with negligible cost. Beyond the empirical validation, we also provide a detailed theoretical analysis of IGPO's time complexity n~\autoref{Budget Analysis}.

\section{Related Work}

The recent success of reinforcement learning (RL) methods in large reasoning models \citep{chen2025surveyinductivereasoninglarge} such as OpenAI o1 \citep{jaech2024openai} and DeepSeek R1 \citep{guo2025deepseek} has established RL as a central tool for enhancing large language models (LLMs)-based agents to solve more complex tasks. A growing body of work has explored different RL algorithms such as PPO \citep{schulman2017proximal}, Reinforce++ \citep{hu2025reinforce++}, GRPO \citep{shao2024deepseekmath}, RLOO \citep{kool2019buy, ahmadian2024back}, DAPO \citep{yu2025dapo}, and GSPO \citep{zheng2025group}. These methods have been particularly effective in improving the capabilities of LLM-based agents \citep{li2025choice}.

Building on these advances, an important line of research has focused on applying RL to search-based agents \citep{deng2025atom, dai2025careful, dai2025evinote}. Early efforts such as DeepRetrieval \citep{jiang2025deepretrieval} demonstrated the feasibility of end-to-end optimization by applying PPO with retrieval-oriented metrics (e.g., recall) as rewards. Subsequent works, including Search-R1 \citep{jin2025search}, DeepResearcher \citep{zheng2025deepresearcher}, and ReSearch \citep{chen2025learning}, extended this paradigm to multi-turn reasoning and search. R1-Searcher \citep{song2025r1} and R1-Searcher++ \citep{song2025r2} further introduced two-stage RL strategies, separately strengthening the ability to interact with retrieval systems and to utilize retrieved information effectively.

However, in multi-turn scenarios, outcome-only rewards remain sparse and often fail to provide sufficient guidance, leading to unstable optimization and inefficient sample utilization. Recent studies have explored step-wise or process-level rewards that assign credit to intermediate actions. ReasonRAG \citep{zhang2025process} adopted Monte Carlo Tree Search (MCTS) to approximate the value of each step. StepSearch \citep{wang2025stepsearch} leveraged a memory vector of retrieved documents, supervising intermediate steps based on their maximum similarity to ground-truth evidence. GiGPO \citep{feng2025group} introduced anchor-based grouping to estimate relative advantages for actions originating from the same anchor state. While these methods provide denser supervision than outcome-only rewards, they either rely on external oracle knowledge or suffer from limited stability and scalability, leaving room for more intrinsic and generalizable process-level reward designs.

\section{Conclusion, Limitations and Future Work}
In this work, we propose IGPO, a simple and effective reinforcement learning framework for training multi-turn LLM-based search agents. By providing intrinsic, ground-truth-aware supervision at every turn while preserving alignment with the final objective, IGPO delivers dense and stable training signals. Extensive experiments across in-domain and out-of-domain benchmarks demonstrate that IGPO consistently outperforms strong baselines, achieving higher accuracy and better sample efficiency, particularly for smaller models where sparse rewards are most problematic. Moreover, IGPO demonstrates remarkable robustness to reward hacking while introducing only negligible computational costs. In doing so, it successfully addresses two critical bottlenecks inherent in existing fine-grained credit assignment RL algorithms: reward hacking and computational inefficiency.

However, our approach still relies on the availability of ground-truth answers, which limits its applicability in open-ended settings. In future work, we plan to extend IGPO to broader agentic scenarios beyond search, including tasks without explicit supervision.

\newpage
\section*{Acknowledgments}
This work was supported by Ant Group Research Intern Program. We thank the Venus Team, Ant Group for their resource support and technical guidance.

\section*{Ethics Statement}
This work follows the ICLR Code of Ethics. No human or animal subjects were involved. All datasets (NQ, TQ, HotpotQA, 2Wiki, MuSiQue, Bamboogle, PopQA) were used according to their respective guidelines, with privacy fully respected. No personally identifiable information was included, and all procedures avoided potential privacy or security risks. Research was conducted with transparency and integrity.

\section*{Reproducibility Statement}
Every effort has been made to ensure the reproducibility of the results reported in this paper. All code and datasets are publicly available in an anonymous repository to facilitate replication and verification. The experimental setup—including training procedures, model configurations, and hardware specifications—is detailed in the appendix to support accurate reproduction of the experiments. These measures are intended to enable other researchers to reproduce the work and contribute to further advancements in the field.

\bibliography{reference}
\bibliographystyle{iclr2026_conference}

\newpage
\appendix

\section{Theoretical Analysis}~\label{appendix:theory}

The theoretical analysis here provides an intuitive support for the efficacy of our proposed method by addressing the limitations of sparse outcome rewards in multi-turn agents.
Specifically, the theory establishes a crucial link: maximizing the process reward (IGPO's objective) is equivalent to minimizing the upper bound on the undesirable accumulation of snowball errors during the reasoning process. 
This minimization, in turn, systematically lowers the theoretical minimum for the final answer error rate, thus providing a fundamental guarantee that IGPO's dense, turn-level signals lead to more confident and successful reasoning trajectories.

\textbf{Notations}.
Let $E_{\text{final}}$ be the event that the agent's generated final answer does not match the ground truth answer. 
Its probability is denoted by $\prob(E_{\text{final}})$, i.e., the error rate.
For each turn $t$ , denote the observed response \texttt{[think]}, \texttt{[tool call]} as $\res_t$. We also posit that there is an unobservable, abstract thinking step $\imp_t$ that underlies the generation of $\res_t$.
Let $\process$ be the process reward, which is a dense reward signal received at each turn of the interaction. It is defined as the information gain about the ground truth answer, which is calculated as the increase in the log-probability of the correct answer from the previous state to the current state.
Then, the total process reward $\total = \sum_{t=1}^{T-1} \mathbb{E}[\process]$ is the cumulative sum of all process rewards over a complete trajectory or episode. The expectation is taken over the thinking step and observed response.
The training objective of the policy is to maximize this total reward.

\begin{definition}[Snowball Error in Multi-turn Agentic RL]
Consistent with~\cite{gan2025rethinking}, we define the information loss at turn $T$ as the conditional entropy $\ent(\imp_t | \res_t)$. 
Consider the non-trivial case where $\left| \ent(\imp_t | \res_t) \right|$ is bounded.
The cumulative snowball error up to turn $T$ is the sum of these losses:
\begin{equation}
    \ent_{<T}(\imp|\res) \triangleq \sum_{t=1}^{T-1} \ent(\imp_t | \res_t)
\end{equation}
\end{definition}

This quantity measures the aggregate uncertainty and ambiguity accumulated throughout the reasoning trajectory before the final answer is produced.

Next, we connect the cumulative snowball error to the agent's final performance. It indicates the fundamental limitation of multi-turn agentic RL pipeline caused by snowball error.

\begin{lemma}[Lower bound of error rate]\label{thm:error_impact}
The probability of a final answer error, $\prob(E_{\text{final}})$, is lower-bounded by the cumulative snowball error accumulated during the reasoning process:
\begin{equation}
    \prob(E_{\text{final}}) = \Omega \left( \frac{\ent_{<T}(\imp|\res)}{T-1} \right) - C_{\text{const}}.
\end{equation}
where $C_{\text{const}}$ is a small positive constant.
\end{lemma}
\begin{proof}[Proof Sketch]
This result is strongly motivated by Theorem 3.3 from~\cite{gan2025rethinking}. We treat the generation of the final answer at turn $T$ as the final step of a multi-step reasoning process. The quality of this final step is conditioned on the information accumulated over the previous $T-1$ turns. The theorem from (Gan et al., 2025) states that the error probability of any step is lower-bounded by the average snowball error accumulated up to that point. Applying this principle to the final step ($t=T$) yields the stated result.
\end{proof}

\begin{assumption}[Monotonic Reward-Information Loss Link] \label{assump:reward_loss}
The expected process reward at any turn $H$, $\mathbb{E}[\process]$, is monotonically non-increasing with respect to the information loss at that turn, $\ent(\imp_t|\res_t)$.
We assume there exists a bounded and monotonically non-increasing convex function $f: \mathbb{R}_+ \to \mathbb{R}$ such that:
\begin{equation}
    \mathbb{E}[\process | \imp_t, \res_t] \le f \left( \ent(\imp_t|\res_t) \right).
\end{equation}

\end{assumption}

\begin{remark}
As the information loss $\ent(\imp_t|\res_t)$ at turn $t$ increases, the expected total information loss tends to decreases and asymptotically approaches a relatively small value, which is characterized by the convex nature of function $f$.

\end{remark}

This assumption leads to the following result, demonstrating that optimizing for process rewards implicitly constrains the accumulation of snowball errors.
We first formalize the intuition that a clearer reasoning step (lower information loss) is a prerequisite for a high-quality query, which in turn yields a higher expected process reward.

\begin{theorem}[Process Reward as a Bound on Snowball Error]\label{thm:reward_bound}
% Let $\pi_{\text{proc}}$ be a policy trained to maximize the total expected process reward $\total$.
Under Assumption \ref{assump:reward_loss}, the expected cumulative snowball error is upper bounded by
\begin{align}
    % \mathbb{E}_{\tau \sim \pi_{\text{proc}}}[\ent_{<H}(\imp|\res)] \le \frac{(H-1)C_{\max} - \total}{\beta}
    \mathbb{E}[\ent_{<T}(\imp|\res)] = \mathcal{O}(1) - \Omega \left(\total\right).
\end{align}
\end{theorem}

Theorem \ref{thm:reward_bound} establishes that maximizing the process reward is mathematically coupled with minimizing an upper bound on the cumulative snowball error.
The combination of Theorem \ref{thm:reward_bound} and Lemma \ref{thm:error_impact} provides a complete, end-to-end theoretical justification for the efficacy of our proposed process reward mechanism. The logical chain is as follows:
\begin{itemize} [leftmargin=0.5cm]
    \item \textbf{Maximizing the process reward} (our algorithm's objective) forces the agent to \textbf{minimize an upper bound on the cumulative snowball error} (Theorem \ref{thm:reward_bound}).
    \item Minimizing the cumulative snowball error, in turn, \textbf{lowers the theoretical minimum for the final error rate}, thereby systematically increasing the probability of task success (Lemma \ref{thm:error_impact}).
\end{itemize}
In conclusion, the turn-level process reward is not merely an engineering heuristic; it is a theoretically grounded mechanism that fundamentally addresses the problem of error accumulation in multi-step reasoning. By providing a dense, immediate signal for reasoning clarity, it transforms the intractable problem of sparse-reward, long-horizon exploration into a series of manageable, short-horizon sub-problems, each aimed at maximizing immediate information gain. This explains the significant gains in training efficiency and final performance observed in our experiments.

\section{Proof for Theoretical Analysis} \label{appendix:proof}

\subsection{Proof of Lemma~\ref{thm:error_impact}}

\begin{proof}
We achieve this by applying Theorem 3.3 from \cite{gan2025rethinking} to the final decision-making step of the agent.
In particular,
\begin{equation}
    \prob(E_{\text{final}}) \ge \frac{\frac{\ent_{<T}(\imp|\res)}{T-1} - C_{1}}{\log(|\mathcal{A}_{\text{final}}|-1)},
\end{equation}
where $|\mathcal{A}_{\text{final}}|$ is the cardinality of the final answer space and $C_{1}$ is a small positive constant analogous to $\ent_b(e_t)$ in \cite{gan2025rethinking}.
Since $\log(|\mathcal{A}_{\text{final}}|-1)$ and $C_{1}$ are constant, $\frac{\frac{\ent_{<T}(\imp|\res)}{T-1} - C_{1}}{\log(|\mathcal{A}_{\text{final}}|-1)}$ simplifies to a form that is asymptotically dominated by the variable term.
Therefore, the right-hand side of the inequality can be expressed in terms of the lower bound symbol $\Omega$ as $\Omega \left( \frac{\ent_{<T}(\imp|\res)}{T-1} \right) - C_{\text{const}}$, which completes the proof.
\end{proof}

\subsection{Proof of Theorem~\ref{thm:reward_bound}}

\begin{proof}

According to the nature of $f$ and the fact that there exist constants $C_{\max}$ and $\beta$ such that for all non-negative bounded $x$, there holds $f(x) \le C_{\max} - \beta x$.
Therefore, by taking the expectation over Assumption \ref{assump:reward_loss} and summing across all turns from $t=1$ to $T-1$, we have
\begin{align*}
    \total = \sum_{t=1}^{T-1} \mathbb{E}[\process] & \le \sum_{t=1}^{T-1} \mathbb{E}[f \left( \ent(\imp_t|\res_t) \right)] 
    \\ & \le \sum_{t=1}^{T-1} \mathbb{E}[C_{\max} - \beta \cdot \ent(\imp_t|\res_t)] 
    \\ & = (T-1)C_{\max} - \beta \sum_{t=1}^{T-1} \mathbb{E}[\ent(\imp_t|\res_t)]
    \\ & = (T-1)C_{\max} - \beta \ \mathbb{E}[\ent_{<T}(\imp|\res)].
\end{align*}
Rearranging terms yields the final result.
\end{proof}

\section{More Implementation Details}
\label{A: training details}

All our training experiments are conducted on 8 × NVIDIA A100-80G GPUs. The detailed hyperparameter settings are provided in ~\autoref{tab:hyperparameters}. Unless otherwise specified, all experiments are based on this configuration.

\begin{table}[H]
\centering
\caption{Training hyperparameters.}
\label{tab:hyperparameters}
\begin{tabular}{cc}
\hline
Training hyperparameters   & Value \\ \hline
Training Batch Size        & 32    \\
Mini-Batch Size            & 512   \\
Infer Tensor Model Parallel Size & 1     \\
Sequence Parallel Size & 4 \\
Max Prompt Length          & 30767 \\
Max Response Length        & 2000  \\
Actor Learning Rate        & 1e-6  \\
Rollout Temperature        & 1.0   \\
Rollout Group Size         & 16    \\
Max Turn Call Turns        & 10    \\ 
KL-Divergence loss coefficient        & 0.001    \\ \hline
\end{tabular}
\end{table}

\begin{table}[htbp]
\centering
\caption{Performance comparison across different IG bases and normalization strategies. We compare Prob-based IG (Joint \& Separate normalization) against LogProb-based IG (Separate normalization). The combination of LogProb and Separate normalization outperforms other settings.}
\label{tab:ablation2}
\resizebox{\columnwidth}{!}{%
\begin{tabular}{@{}lccccccccc@{}}
\toprule
\multicolumn{1}{c}{} & \multicolumn{4}{c}{In-domain}                                 & \multicolumn{1}{l}{} & \multicolumn{3}{c}{Out-of-domain}             & \multicolumn{1}{l}{}     \\ \cmidrule(lr){2-5} \cmidrule(lr){7-9}
Method               & NQ            & TQ            & HotpotQA      & 2Wiki         & \multicolumn{1}{l}{} & Musique       & Bamboogle     & PopQA         & \multicolumn{1}{l}{Avg.} \\ \midrule
\multicolumn{10}{l}{\cellcolor[HTML]{EFEFEF}\textbf{Qwen2.5-3B-Instruct}}                                                                                                              \\
Prob+Joint                & 40.5    & \textbf{69.4}    & 46.8          & 48.2          &                      & 23.1          & {\ul 57.9}          & 47.4    & 47.6                     \\
Prob+Separate          & {\ul41.2}          & 68.9          & {\ul 47.2}    & {\ul 49.5}    &                      & {\ul 23.5}    & 57.7    & {\ul 48.3}          & {\ul 48.0}               \\
Logprob+Separate        & \textbf{41.9} & {\ul 69.2} & \textbf{47.8} & \textbf{51.4} & \textbf{}            & \textbf{24.8} & \textbf{58.4} & \textbf{49.0} & \textbf{48.9}            \\ \midrule
\multicolumn{10}{l}{\cellcolor[HTML]{EFEFEF}\textbf{Qwen2.5-7B-Instruct}}                                                                                                              \\
Prob+Joint                & \textbf{46.7}    & 80.1    & 57.2          & 68.2          &                      & 31.4          & {\ul 74.9}          & 52.5    & 58.7                     \\
Prob+Separate          & 46.2          & {\ul 80.3}          & {\ul 58.2}    & {\ul 71.4}    &                      & {\ul 31.8}    & 74.6    & {\ul 53.6}          & {\ul 59.4}               \\
Logprob+Separate        & {\ul 46.4} & \textbf{80.6} & \textbf{59.0} & \textbf{72.1} & \textbf{}           & \textbf{32.7} & \textbf{77.0} & \textbf{53.8} & \textbf{60.2}            \\ \bottomrule
\end{tabular}
}

\end{table}

\section{More Discussion and Experimental Analysis}
\label{app:case study and more discuss}
\subsection{Comparison of Info. Gain Basis (Prob. vs. LogProb) and Normalization Strategies (Joint vs. Separate)}
We investigate the impact of different information gain computation bases (probability vs. log-probability) and normalization strategies: joint (normalizing all rewards collectively) vs. separate (normalizing IG and outcome independently). Note that we exclude the Logprob+Joint combination due to the significant scale disparity between log-based IG and bounded outcome rewards, which renders joint normalization ineffective. As shown in~\autoref{tab:ablation2}, the combination of log-probability-based information gain and separate normalization (Logprob+Separate) emerges as the optimal strategy. Specifically, switching from joint to separate normalization (Prob+Joint → Prob+Separate) yields a clear gain (+0.7 on 7B Avg, +0.7 on 3B Avg), validating the necessity of decoupling the statistics of intermediate and final rewards. Replacing probability with log-probability (Prob+Separate → Logprob+Separate) provides an additional boost (+0.8 on 7B Avg, +0.9 on 3B Avg), demonstrating the numerical stability advantages of log-probability. These improvements are consistent across both model scales and domain types, demonstrating the robustness of our design choices.
\label{info_gain&norm}
\subsection{Comparison with Other Process-Reward Methods}
In addition to its obvious performance advantages, we also conduct a deeper analysis of IGPO's superiority in terms of algorithmic characteristics compared to other process-reward-based agentic RL algorithms. We first introduce other existing process-reward-based agentic RL algorithms:
\begin{itemize} [leftmargin=0.5cm]
    \item {\textbf{ReasoningRAG}}. The main contribution of this work is the proposal of a step-level data labeling strategy based on MCTS. Subsequently, the DPO algorithm is used to optimize the agent’s policy on the labeled step-level dataset. The main limitations of this method are: (1) the data labeling process relies on MCTS, which is inefficient, and when the number of samples is insufficient, it is difficult to accurately estimate the value of each step; (2) the off-policy optimization based on DPO is less effective than on-policy algorithms.
    \item {\textbf{StepSearch}}. StepSearch constructs turn-level supervision signals by pre-defining golden search keywords and golden tool responses, and adopts an on-policy optimization approach. Although it shifts from off-policy to on-policy, the annotation process is resource-intensive and prone to annotator bias (whether from humans or LLMs).
    \item {\textbf{GiGPO}}. GiGPO introduces a step-level grouping strategy based on anchor states and performs fine-grained advantage estimation within each step-level group. Although this provides a novel solution, it essentially still relies on the Monte Carlo assumption. When the number of anchor states is insufficient, it is often difficult to accurately estimate their value, which in turn leads to biased advantage estimation.
\end{itemize}
The proposed IGPO effectively addresses the aforementioned limitations. Starting from the on-policy GRPO setting (where rollout data are used for a single parameter update), it employs an information-gain–based incremental reward construction strategy that requires no annotation and does not rely on Monte Carlo. Moreover, the incorporation of ground-truth awareness substantially reduces bias. \autoref{tab: various process reward} provides a detailed comparison highlighting the advantages of IGPO over other algorithms.

\begin{table}[H]
\setlength{\tabcolsep}{1.2mm}
\caption{Comparison between various process reward methods.}
\label{tab: various process reward}
\begin{tabular}{lcccc}
\hline
Algorithm    & On-Policy & Explicit Labeling-Free & Monte Carlo–Free & Introduces No Bias    \\ \hline
ReasoningRAG & No        & Yes                    & No               & Sample-size Dependent \\
StepSearch   & Yes       & No                     & Yes              & No                    \\
GiGPO        & Yes       & Yes                    & No               & Sample-size Dependent \\
IGPO         & Yes       & Yes                    & Yes              & Yes                   \\ \hline
\end{tabular}
\end{table}

\subsection{Time breakdown of each stage in IGPO training (same as GRPO)}
\label{time_breakdown}

\begin{table}[H]
\centering
\caption{We have calculated the average time percentage spent on each phase of a training step for IGPO (same as GRPO). The majority of the time is spent on Sampling (Rollout), with Recompute Log-Prob accounting for less than 5\% of the total duration. The time spent on the Return/Advantage Computation phase is much smaller than 1\% and can be ignored.}
\label{tab:time breakdown}
\begin{tabular}{@{}ccccc@{}}
\toprule
Phase & Sampling & Param Update & Recompute Log-Prob & Return/Advantage Comp.    \\ \midrule
Time Proportion &  82.6\%   & 12.6\%           & 4.8\%                       & \textless{}\textless 1\% \\ \bottomrule
\end{tabular}
\end{table}

\subsection{Spurious Correlations Analysis}
\label{spurious correlations}
\begin{figure}[htbp] 
    \centering
    \includegraphics[width=1\textwidth]{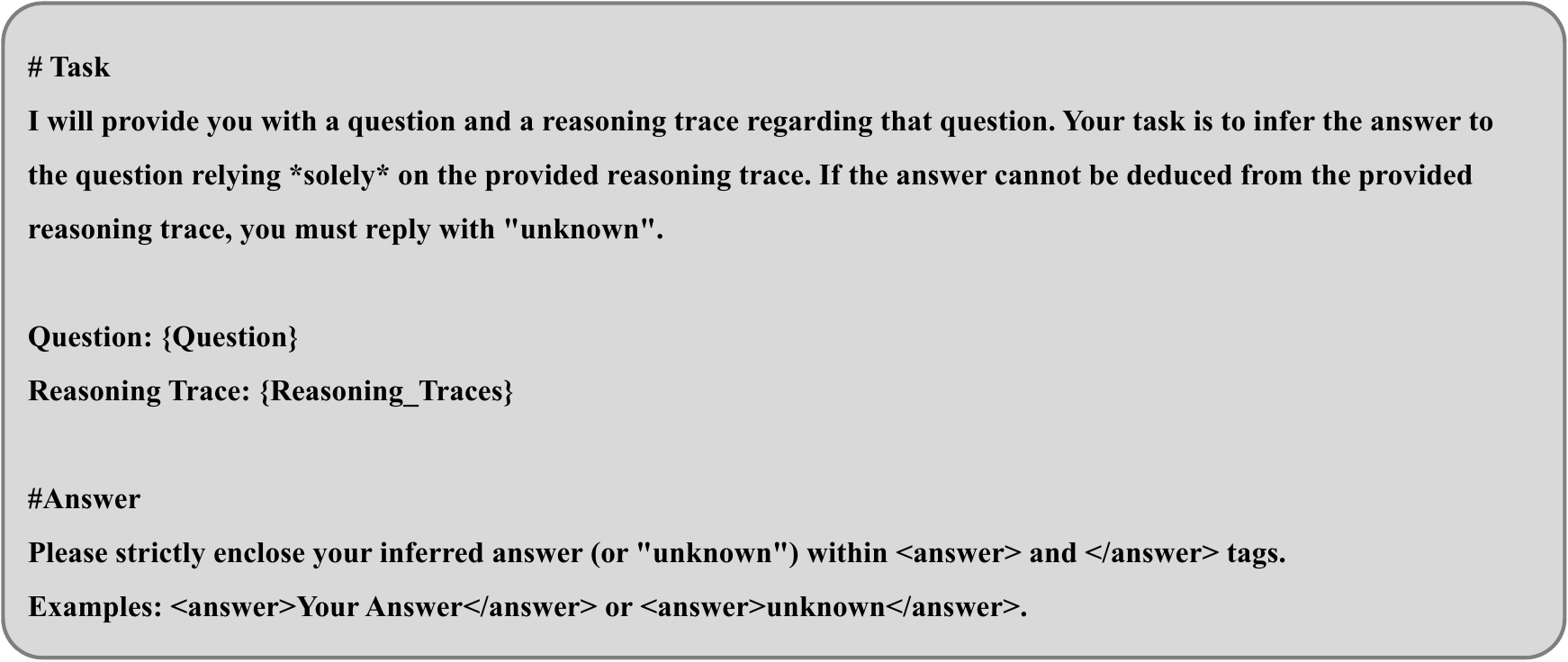}
    \caption{Prompt template for gemini-2.5-pro in the spurious correlations analysis.}
    \label{fig:judge_prompt}
\end{figure}

It is widely acknowledged that LLMs often exploit spurious correlations to solve problems—achieving correct answers through unfaithful or erroneous intermediate reasoning—rather than learning the genuine underlying reasoning process. This tendency significantly compromises their out-of-distribution performance. To investigate whether the proposed IGPO mitigates or exacerbates such spurious correlations, we conduct the following experiment:

\paragraph{Experiment Setup.}We select the set of test samples correctly answered (F1=1.0) by both IGPO and GRPO agents. We extract the corresponding outputs, remove the final answers, and retaine only the intermediate reasoning traces. Subsequently, we employe a powerful teacher LLM (gemini-2.5-pro~\citep{comanici2025gemini}) to deduce the final answer based on these reasoning paths. By comparing the accuracy of the answers inferred from the IGPO versus GRPO reasoning traces, we assesse whether IGPO is more prone to yielding 'correct answers with incorrect or unfaithful reasoning' (i.e., spurious correlations). The prompt template used for gemini-2.5-pro is illustrated in~\autoref{fig:judge_prompt}.
\paragraph{Result.}As shown in \autoref{tab:spurious correlations}, for both 3B and 7B models, the accuracy of the teacher model in inferring answers from IGPO-agent traces consistently outperforms that from GRPO-agent traces across all seven datasets, including both in-domain and out-of-domain tasks. This indicates that the reasoning traces generated by the IGPO agent are more informative and of higher quality. It demonstrates that IGPO effectively mitigates spurious correlations through ground-truth guided fine-grained credit assignment, further validating its generalization capabilities.
\paragraph{Additional Evidence.}Beyond the aforementioned experiment, we provide the following evidence to further support that IGPO mitigates spurious correlations: (i) \textbf{Superior OOD Performance}. According to the results in \autoref{tab:ablation}, compared to GRPO, IGPO achieves an average performance gain of 12.6\% (7B) and 42.8\% (3B) on In-domain datasets (NQ, TQ, HotpotQA, and 2Wiki), whereas on Out-of-domain datasets (Musique, Bamboogle, PopQA), the average improvement increases to 13.7\% (7B) and 55.2\% (3B). The fact that performance gains in OOD settings exceed those in ID settings contradicts the pattern of spurious correlations, which typically favors ID performance at the expense of OOD generalization. (ii) \textbf{Exceptional Multi-hop Capabilities}. As indicated in \autoref{tab:ablation}, IGPO outperforms GRPO with an average improvement of 7.4\% (7B) and 29.5\% (3B) on single-hop tasks (NQ, TQ, PopQA), while on multi-hop tasks (HotpotQA, 2Wiki, Musique, Bamboogle), the average improvement reaches 17.8\% (7B) and 68.1\% (3B). The performance boost on multi-hop tasks is significantly greater than on single-hop tasks. This is also inconsistent with spurious correlation patterns, which are prone to appearing in multi-hop scenarios and consequently causing greater detriment to performance. Therefore, the superior performance in both OOD and multi-hop scenarios serves as further evidence that IGPO effectively mitigates spurious correlations.
\begin{table}[htbp]
\centering
\caption{Results of spurious correlations analysis. We select test samples where both IGPO and GRPO agents achieve correct answers (F1=1.0). Using gemini-2.5-pro, we infer answers solely based on the reasoning traces from these samples to compare the informativeness of the traces generated by each method ("All" denotes the aggregated accuracy across all test samples). The results demonstrate that, compared to GRPO, IGPO yields reasoning traces that are more informative and of higher quality, further validating its generalization capabilities.}
\label{tab:spurious correlations}
\resizebox{\columnwidth}{!}{%
\begin{tabular}{@{}lccccccccc@{}}
\toprule
\multicolumn{1}{c}{} & \multicolumn{4}{c}{In-domain}                                 & \multicolumn{1}{l}{} & \multicolumn{3}{c}{Out-of-domain}             & \multicolumn{1}{l}{}     \\ \cmidrule(lr){2-5} \cmidrule(lr){7-9}
Method               & NQ            & TQ            & HotpotQA      & 2Wiki         & \multicolumn{1}{l}{} & Musique       & Bamboogle     & PopQA         & \multicolumn{1}{l}{All} \\ \midrule
\multicolumn{10}{l}{\cellcolor[HTML]{EFEFEF}\textbf{Qwen2.5-3B-Instruct}}                                                                                                              \\
GRPO                & 93.1    & 89.0    & 90.1          & 94.7          &                      & 84.3          & 89.7          & 93.5    & 91.2                     \\
IGPO           & \textbf{95.4}          & \textbf{92.8}          & \textbf{94.2}    & \textbf{97.7}    &                      & \textbf{90.2}    & \textbf{92.3}    & \textbf{96.1}          & \textbf{94.6}               \\
             \midrule
\multicolumn{10}{l}{\cellcolor[HTML]{EFEFEF}\textbf{Qwen2.5-7B-Instruct}}                                                                                                              \\
GRPO                & 86.5    & 88.8    & 85.8          & 92.0          &                      & 76.1          & 91.2          & 95.1    & 89.0                     \\
IGPO          & \textbf{89.5}          & \textbf{91.1}          & \textbf{92.6}    & \textbf{94.4}    &                      & \textbf{83.0}    & \textbf{94.1}    & \textbf{95.1}          & \textbf{92.0}               \\
            \bottomrule
\end{tabular}%
}
\end{table}

\section{Failure Analysis}
\label{failure}
Despite IGPO's superior performance, to ensure a comprehensive analysis, we investigated its failure modes, specifically examining instances where IGPO exhibits performance degradation (i.e., F1 scores lower than GRPO). As shown in \autoref{tab:failure}, minor degradation is observed across datasets, with IGPO underperforming GRPO on approximately 3.6\% of the test samples overall. While this degradation is marginal, it warrants in-depth analysis.

Algorithmically, IGPO extends GRPO by computing the log probability increment of the ground truth answer between adjacent turns. This constructs a turn-level reward, providing stronger, denser, and ground-truth-aware supervision for fine-grained guidance. While GRPO relies solely on final outcome correctness, both methods depend on ground truth quality. Consequently, while IGPO amplifies the benefits of high-quality data, it inevitably exacerbates the impact of noise within the ground truth.

\begin{figure}[H]
    \centering
    \includegraphics[width=1\textwidth]{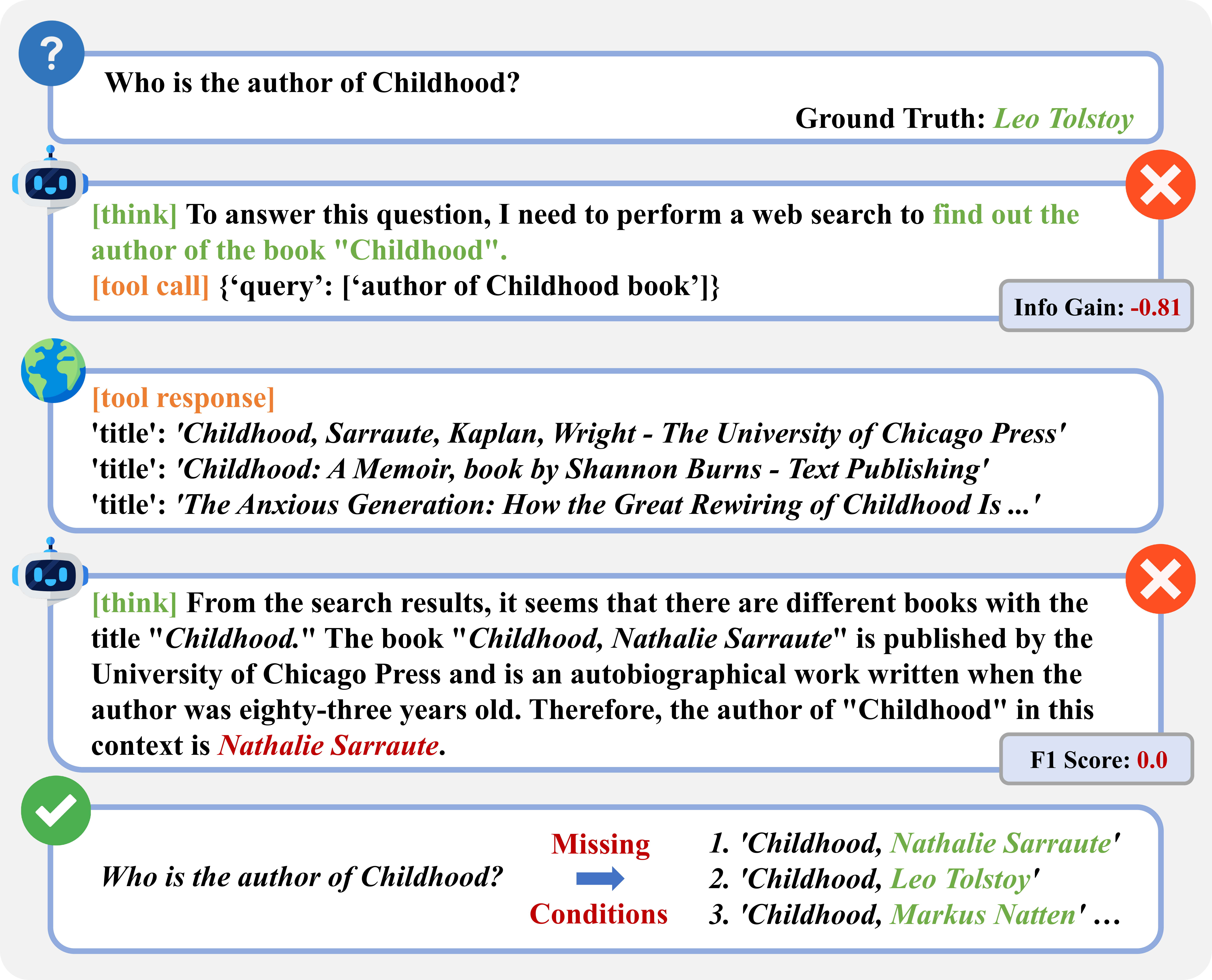}
    
    \caption{The query ‘Who is the author of Childhood?’ (a question from HotpotQA) is inherently ambiguous because multiple unrelated works share the same title—e.g., Childhood by Leo Tolstoy (fiction, ground truth), Nathalie Sarraute (autobiography, the model’s answer), and Markus Natten (poem). Without specifying which literary work is intended, several factually correct answers exist. Consequently, when the model outputs a correct but non–ground-truth author, it is penalized as ‘wrong,’ producing misleading negative rewards. Such mislabeled supervision degrades the effectiveness of IGPO by punishing valid reasoning aligned with alternative correct interpretations}.
    \label{fig:failure}
\end{figure}

Through detailed data inspection, we identified a representative pattern of ground truth failure: ambiguous questions lacking specific conditions, leading to multiple valid answers. In such scenarios, IGPO is prone to degradation. If the model leverages reasoning and tool usage to increase the probability of a factually correct—but non-ground-truth—answer, it incurs a penalty from the turn-level reward. This erroneously suppresses valid behaviors and impairs the model's reasoning capabilities. \autoref{fig:failure} illustrates a real training instance from HotpotQA. The question, "Who is the author of Childhood?", lacks context (e.g., genre), allowing for multiple valid answers. When the model retrieved information regarding "Nathalie Sarraute" (a correct answer, though not the designated ground truth), it was penalized (Info Gain = -0.81). This constitutes a false suppression of correct reasoning and tool usage. While GRPO also struggles with such ambiguity, IGPO amplifies this noise, resulting in a more significant negative impact. We observed frequent occurrences of this pattern in the training set and identify it as the primary cause of IGPO's performance degradation.

It is important to note, however, that this degradation stems from data defects rather than algorithmic flaws. The ability of IGPO to maintain high performance despite these data imperfections (with a degradation rate of only 3.6\%) effectively demonstrates its robustness. We look forward to exploring more complex failure modes in future work.

\begin{table}[htbp]
\centering
\caption{We compared the F1 scores of IGPO and GRPO on the test set and analyzed the proportion of samples falling into three categories: IGPO $>$ GRPO, IGPO $=$ GRPO, and IGPO $<$ GRPO. Overall, IGPO exhibits slight performance degradation across datasets, with approximately 3.6\% of test samples showing lower performance compared to GRPO.}
\label{tab:failure}
\begin{tabular}{@{}lccc@{}}
\toprule
Dataset   & IGPO\textgreater{}GRPO & IGPO=GRPO & IGPO\textless{}GRPO \\ \midrule
2Wiki     & 35.8\%                 & 59.6\%    & 4.6\%               \\
Bamboogle & 47.2\%                 & 49.6\%    & 3.2\%               \\
HotpotQA  & 49.2\%                 & 48.4\%    & 2.4\%               \\
Musique   & 71.2\%                 & 25.4\%    & 3.4\%               \\
NQ        & 57.4\%                 & 40.4\%    & 2.2\%               \\
PopQA     & 42.8\%                 & 53.4\%    & 3.8\%               \\
TQ        & 33.6\%                 & 61.2\%    & 5.2\%               \\
All       & 48.3\%                 & 48.1\%    & 3.6\%               \\ \bottomrule
\end{tabular}
\end{table}

\section{Computation Budget Theoretical Analysis}
\label{Budget Analysis}
This section will provide a detailed analysis of the additional FLOPs introduced by IGPO. Since the extra computation introduced by IGPO occurs during the \texttt{recompute\_log\_prob} phase (a single forward propagation), we will analyze the FLOPs of a single forward propagation in the Transformer model to examine the additional computation introduced by IGPO.
\subsection{FLOPs calculation}
\label{FLOPs}
FLOPs (Float Point operations) represents the number of floating-point operations, commonly used to measure the computational complexity. We will focus on the FLOPs calculation in matrix multiplication. For matrices $A\in R^{1\times n}$ and $B\in R^{n\times 1}$, computing $AB$ requires n multiplication operations and n addition operations, totaling $2n$ FLOPs. Therefore, the FLOPs required to compute $AB$ for matrices $A\in R^{m\times n}$ and $B\in R^{n\times p}$ is $2nmp$.
\subsection{symbols}
Let 
$b$: batch\_size,
$s$: sequence length,
$g$: ground-truth answers length,
$h$: hidden state dimension (assume the intermediate dimension = $4h$),
$head_{num}$: number of attention heads,
$head_{dim}$: dimension of attention head,       
$l$: number of layers,
$\mathcal{V}$: vocabulary size,
$x$: the input data,
$W_Q$: the query matrix,
$W_K$: the key matrix,
$W_V$: the value matrix,
$W_o$: the output matrix of the attention module,
$W_{up}$: the up-projection matrix in MLP module,
$W_{down}$: the down-projection matrix in MLP module,
$y_{attn}$: the output of attention module,
$y_{mlp}$: the output of MLP module,
$\sigma$: the activation function. 

\subsection{Additional FLOPs introduced by IGPO compared to GRPO.}
We first analyze the FLOPs of GRPO's \texttt{recompute\_log\_prob} phase, which refers to the FLOPs of a single forward propagation. The shape of the input data is $[b,s]$. We first analyze the self-attention module, whose computation process is as follows:
\begin{equation}
\label{attention1}
    Q=xW_Q,\quad K=xW_K, \quad V=xW_{V},
\end{equation}
\begin{equation}
\label{attention2}
    y_{attn}=\text{softmax}(\frac{QK^T}{\sqrt{h}})\cdot V\cdot W_o + x.
\end{equation}
The input and output shapes of the matrix multiplication in Eq.~\ref{attention1} are: $[b,s,h]\times [h,h]\rightarrow [b,s,h]$. Based on ref~{FLOPs}, the FLOPs for this process is:
\begin{equation}
    FLOPs_{attn1}=3\times 2bsh^2=6bsh^2.
\end{equation}
For $QK^T$ in Eq.~\ref{attention2} the input and output shapes of the matrix multiplication are: $[b,head_{num},s,head_{dim}]\times [b,head_{num}, head_{dim},s]\rightarrow [b,head_{num},s,s]$, the FLOPs for this process is $2bs^2h$.

For \texttt{attention\_score}$\cdot V$ in Eq.~\ref{attention2} the input and output shapes of the matrix multiplication are: $[b,head_{num},s,s]\times [b,head_{num},s,  head_{dim}]\rightarrow [b,head_{num},s,head_{dim}]$, the FLOPs for this process is $2bs^2h$.

For the linear mapping operation of multiplying by $W_o$ in Eq.~\ref{attention2}, the input and output shapes of the matrix multiplication are: $[b,s,h]\times [h,h]\rightarrow [b,s,h]$, the FLOPs for this process is $2bsh^2$.

Therefore, the total FLOPs in Eq.~\ref{attention2} are:
\begin{equation}
    FLOPs_{attn2}=2bs^2h+2bs^2h+2bsh^2=4bs^2h+2bsh^2.
\end{equation}

Next, we analyze the FLOPs of the MLP module. We assume the intermediate dimension (i.e., the up-projection dimension) of the MLP module is $4h$, which is a common setting. The computation process of the MLP module can be expressed as follows:
\begin{equation}
\label{mlp}
    y_{mlp}=\sigma(y_{attn}W_{up})W_{down}+y_{attn}.
\end{equation}
For the up-projection operation in Eq.~\ref{mlp}, the input and output shapes of the matrix multiplication are: $[b,s,h]\times [h,4h]\rightarrow [b,s,4h]$ the FLOPs for this process is $8bsh^2$. And, for the down-projection operation, the input and output shapes of the matrix multiplication are: $[b,s,4h]\times [4h,h]\rightarrow [b,s,h]$ the FLOPs for this process is $8bsh^2$. Therefore, the total FLOPs of the MLP model are:
\begin{equation}
    FLOPs_{mlp}=2\times 8bsh^2=16bsh^2.
\end{equation}

At this point, we have calculated the FLOPs required for a Transformer module during the forward propagation process: $FLOPs_{attn1}+FLOPs_{attn2}+FLOPs_{mlp}$. Therefore, the FLOPs for the entire Transformer model during the forward propagation process are: $l\times(FLOPs_{attn1}+FLOPs_{attn2}+FLOPs_{mlp})$.

In addition, due to the large size of the vocabulary $\mathcal{V}$, the computation involved in mapping the hidden state (dimension = $h$) to logits (dimension = $\mathcal{V}$) at the end is also significant and cannot be ignored. The input and output shapes of the matrix multiplication are: $[b,s,h]\times[h,\mathcal{V}]$, the FLOPs for this process is $FLOPs_{logits}=2bsh\mathcal{V}$.

Therefore, the FLOPs of the GRPO's \texttt{recompute\_log\_prob} phase are:
\begin{equation}
    \begin{aligned}
    FLOPs_{GRPO}&=l\times(FLOPs_{attn1}+FLOPs_{attn2}+FLOPs_{mlp})+FLOPs_{logits}
    \\ & = 24lbsh^2+4lbs^2h+2bsh\mathcal{V}
    \end{aligned}.
\end{equation}

For the sake of simplicity in the subsequent analysis, we let $\alpha=4lbh,\beta=24lbh^2 + 2bh\mathcal{V}$, The FLOPs of GRPO can be expressed as:
\begin{equation}
    FLOPs_{GRPO}=\alpha s^2 + \beta s.
\end{equation}

Since we integrated the process of calculating the ground truth answer's log probability into the \texttt{recompute\_log\_prob} phase by extending the attention mask matrix~\ref{cost}, the FLOPs for the IGPO's \texttt{recompute\_log\_prob} phase are:
\begin{align}
    FLOPs_{IGPO}&=\alpha (s+g)^2 + \beta (s+g) \label{FLOPs_IGPO1},
    \\ & \approx \alpha(s^2+2sg) + \beta(s+g), \quad \text{via}\quad g<<s \label{FLOPs_IGPO2}.
\end{align}
Eq.~\ref{FLOPs_IGPO1} to Eq.~\ref{FLOPs_IGPO2} use a first-order Taylor expansion, as $g$ (on the order of $10^1$) is much smaller than $s$ (on the order of $10^4$) during actual training.

The additional FLOPs introduced by IGPO compared to GRPO are:
\begin{equation}
\begin{aligned}
        \Delta FLOPs&=FLOPs_{IGPO}-FLOPs_{GRPO}\\
        &=2\alpha sg+\beta g
\end{aligned}.
\end{equation}
The proportion of additional FLOPs introduced compared to GRPO is:
\begin{equation}
\begin{aligned}
        \frac{\Delta FLOPs}{FLOPs_{\text{GRPO}}} &=\frac{2\alpha sg+\beta g}{\alpha s^2 + \beta s} \\
        &=\frac{g(\alpha s+\beta)+\alpha sg}{s(\alpha s+\beta)} \\
        &=\frac{g}{s}+\frac{\alpha sg}{\alpha s(s+\frac{\beta}{\alpha})} \\
        &=\frac{g}{s} + \frac{g}{s + \frac{\beta}{\alpha}} \\
        &< \frac{2g}{s}
\end{aligned}.
\end{equation}
In actual training scenarios, $\frac{2g}{s}$ is typically on the order of $10^{-3}$, so the additional FLOPs introduced by the IGPO's \texttt{recompute\_log\_prob} phase are only about one-thousandth of the original, and can be considered negligible. Considering the real time proportion of \texttt{recompute\_log\_prob} phase in \autoref{tab:time breakdown}, IGPO introduces only about one ten-thousandth of the additional time consumption compared to GRPO per training step.

\newpage
\section{Comparison between GRPO and IGPO}
\label{appx:grpo vs igpo}
Algorithm~\autoref{alg:grpo-vs-igpo} illustrates the algorithmic flow of IGPO (right) and GRPO (left). The key steps corresponding to each algorithm are highlighted with the same color font to visually highlight the differences: yellow for \textcolor{myYellow}{reward calculation}, green for \textcolor{myGreen}{discounted/advantage estimation}, blue for \textcolor{myBlue}{credit assignment}, and purple for \textcolor{myPurple}{policy optimization}. In terms of reward calculation, IGPO constructs dense turn-level rewards through incremental information gain. For discounted return/advantage estimation, GRPO performs group-wise normalization, whereas IGPO performs group-wise normalization followed by discounted accumulation. Regarding credit assignment, GRPO directly assigns the outcome-based advantage to all tokens of the current output, while IGPO performs turn-level credit assignment. In policy optimization, IGPO achieves efficient and effective optimization by maximizing the turn-level discounted returns.

\begin{algorithm}[htbp]
\caption{\textbf{GRPO} vs. \textbf{IGPO}}
\label{alg:grpo-vs-igpo}

\begin{minipage}[t]{0.49\linewidth}
\textbf{GRPO}
\begin{algorithmic}[1]
\REQUIRE initial policy $\pi_{\theta_{\text{init}}}$; task prompts $\mathcal{D}$; hyperparameters $\epsilon$, $\beta$, $\mu$
\STATE $\pi_{\theta} \gets \pi_{\theta_{\text{init}}}$
\FOR{iteration $=1,\ldots,I$}
  \STATE $\pi_{\text{ref}} \gets \pi_{\theta}$
  \FOR{step $=1,\ldots,M$}
    \STATE Sample a batch $\mathcal{D}_b$ from $\mathcal{D}$
    \STATE $\pi_{\theta_{\text{old}}} \gets \pi_{\theta}$
    \STATE For each $q\in \mathcal{D}_b$, sample $G$ outputs $\{y_i\}_{i=1}^G \sim \pi_{\theta_{\text{old}}}(\cdot \mid q)$
    \textcolor{myYellow}{\STATE Compute outcome reward $\{r_i\}_{i=1}^G$ from the final answer in each $y_i$}
    \textcolor{myGreen}{\STATE Compute each $y_i$'s advantage $\{A_i\}_{i=1}^G$ via group normalization of $\{r_i\}_{i=1}^G$ (Eq.in Sec~\ref{Agentic RL Pipline})}
    \textcolor{myBlue}{\STATE Assign $A_i$ to all tokens of $y_i$}
    \textcolor{myPurple}{\FOR{GRPO iter $=1,\ldots,\mu$}
      \STATE Update $\pi_{\theta}$ by maximizing the GRPO objective (Eq.~\ref{eq:grpo})
    \ENDFOR}
  \ENDFOR
\ENDFOR
\end{algorithmic}
\end{minipage}
\hfill
\begin{minipage}[t]{0.49\linewidth}
\textbf{IGPO}
\begin{algorithmic}[1]
\REQUIRE initial policy $\pi_{\theta_{\text{init}}}$; task prompts $\mathcal{D}$; max turns $H$; hyperparameters $\epsilon$, $\beta$, $\gamma$, $\mu$
\STATE $\pi_{\theta} \gets \pi_{\theta_{\text{init}}}$
\FOR{iteration $=1,\ldots,I$}
  \STATE $\pi_{\text{ref}} \gets \pi_{\theta}$
  \FOR{step $=1,\ldots,M$}
    \STATE Sample a batch $\mathcal{D}_b$ from $\mathcal{D}$
    \STATE $\pi_{\theta_{\text{old}}} \gets \pi_{\theta}$
    \STATE For each $q\in \mathcal{D}_b$, sample $G$ outputs $\{y_i\}_{i=1}^G \sim \pi_{\theta_{\text{old}}}(\cdot \mid q)$
    \textcolor{myYellow}{\FOR{iteration $=1,\ldots,T$}
      \IF{$\text{t} < \text{T}$}
        \STATE Compute the infomation gain-based turn-level rewards $\{r_{i,t}\}_{i=1}^G$ for each $y_i$ (Eq.~\ref{eq:info gain reward})
      \ELSE
        \STATE Compute the final-turn rewards $\{r_{i,T}\}_{i=1}^G$ based on the answer in each $y_i$ (Eq.~\ref{eq:reward})
      \ENDIF
    \ENDFOR}
    \textcolor{myGreen}{\STATE Normalize the turn rewards within the group to derive $\{\tilde{r}_{i,1 \le t \le T}\}_{i=1}^G$ (Eq.~\ref{eq:norm})}
    \textcolor{myGreen}{\STATE Compute the turn-level discounted returns $\{\tilde{R}_{i,1 \le t \le T}\}_{i=1}^G$ in each $y_i$ (Eq.~\ref{eq:disc-adv}).}
    \textcolor{myBlue}{\STATE Assign $\{\tilde{R}_{i,1 \le t \le T}\}_{i=1}^G$ to the tokens in each turn}
    \textcolor{myPurple}{\FOR{IGPO iteration $=1,\ldots,\mu$}
      \STATE Update $\pi_{\theta}$ by maximizing the IGPO objective (Eq.~\ref{eq:igpo})
    \ENDFOR}
  \ENDFOR
\ENDFOR
\end{algorithmic}
\end{minipage}

\end{algorithm}

\newpage
\section{Prompt template used in our experiments.}
\label{app:prompt}
\begin{figure}[htbp] 
    \centering
    \includegraphics[width=1\textwidth]{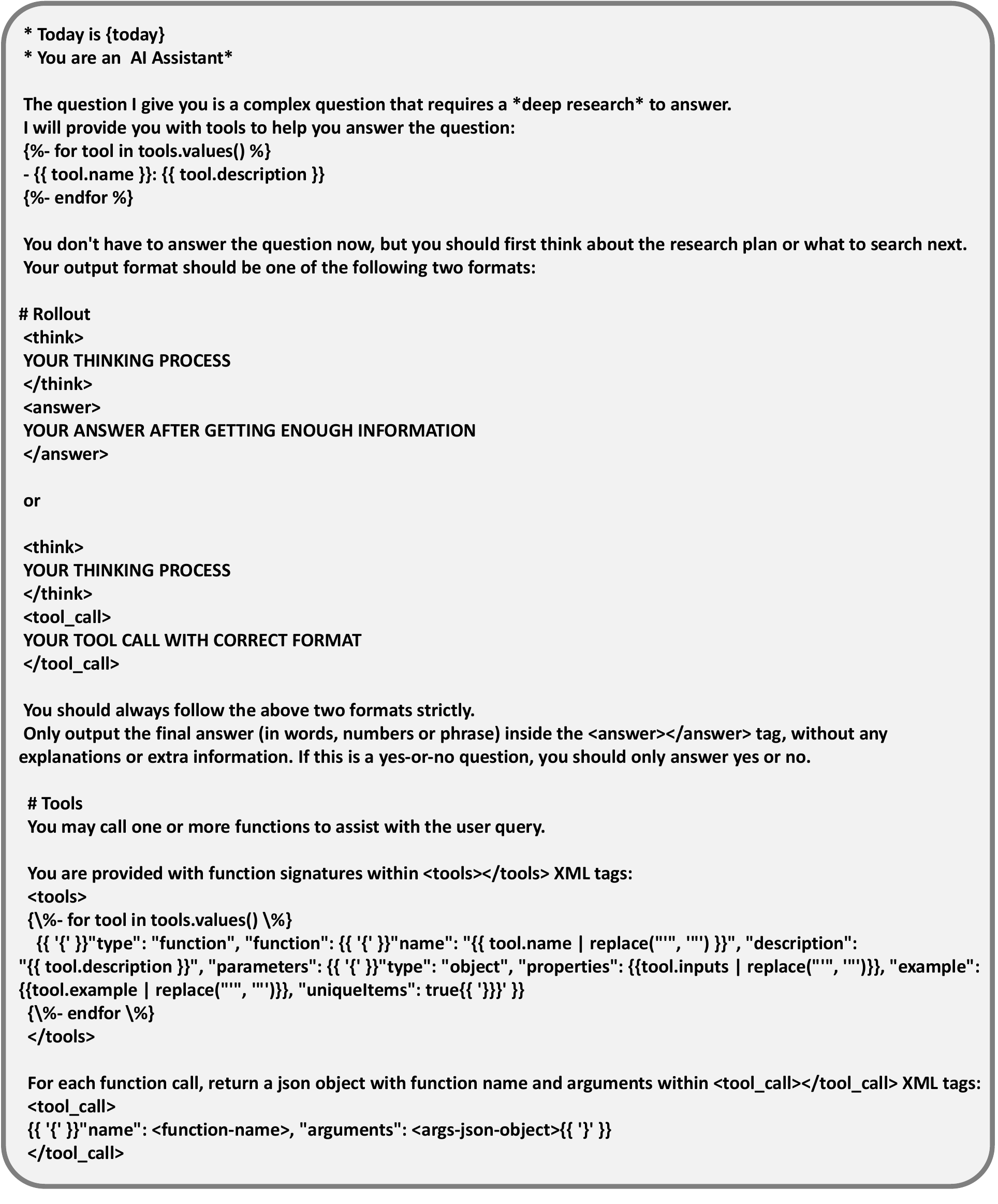}
    \caption{Prompt template used in our experiments.}
    \label{fig:prompt}
\end{figure}

Our prompt follows the style of DeepResearcher~\cite{zheng2025deepresearcher}, and the same template is used for training, validation, and testing. The prompt template is shown in the \autoref{fig:prompt}, where \texttt{\{today\}} represents the current date to ensure the relevance of the model’s response. \texttt{\{\{ tool.name \}\}: \{\{ tool.description \}\}} indicates the available tools, while the \texttt{\#Rollout section} controls the model’s output format. The \texttt{\#Tools} section provides the model with the tool invocation method.

\newpage
\section{Case Study}
\label{case study}
\begin{figure}[htbp]
    \centering
    \includegraphics[width=1\textwidth]{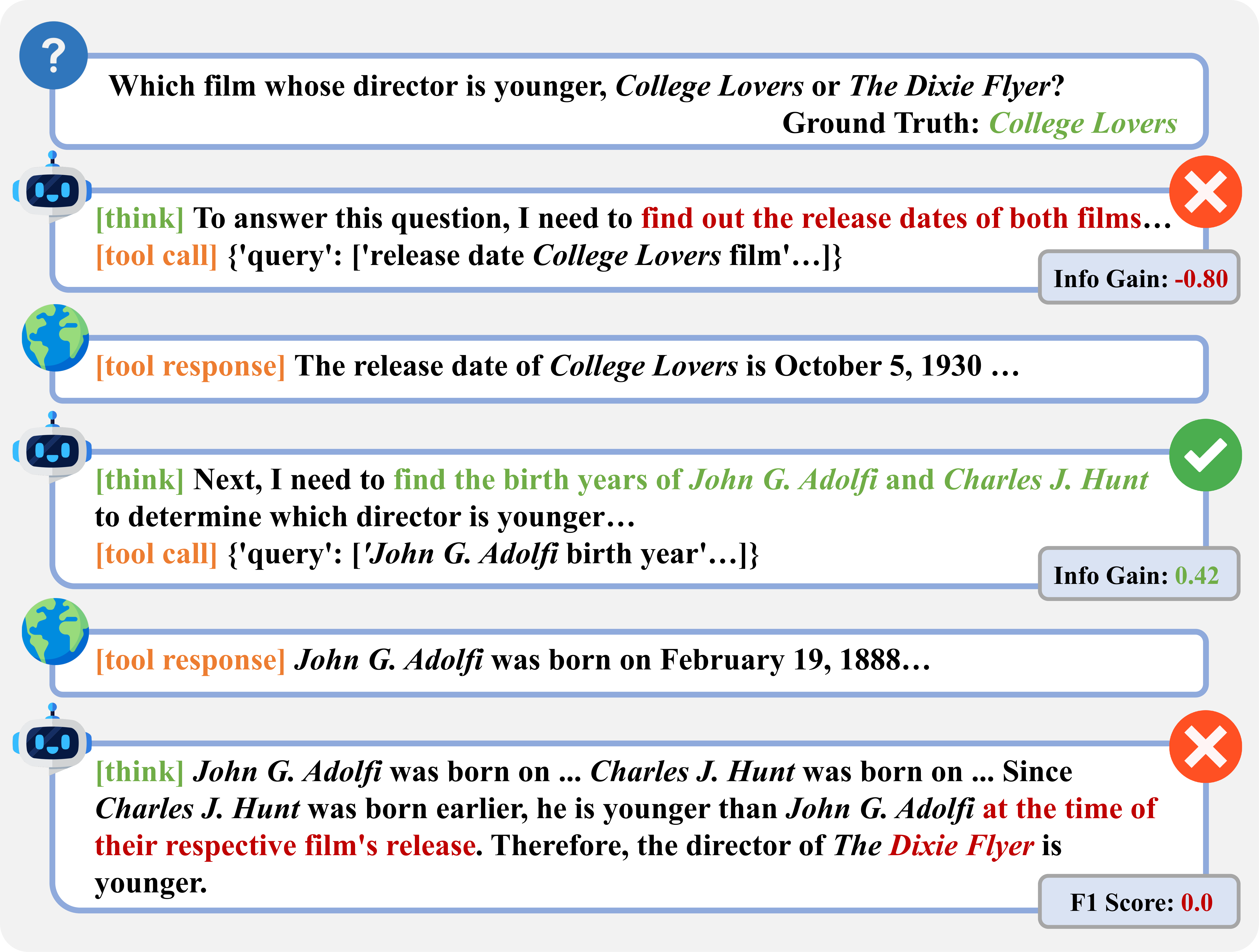}
    \caption{Case study showing a scenario where the final answer is incorrect but contains a single correct retrieval turn. IGPO provides a process reward for this turn, improving token utilization.}
    \label{case1}
\end{figure}

\begin{figure}[htbp]
    \centering
    \includegraphics[width=1\textwidth]{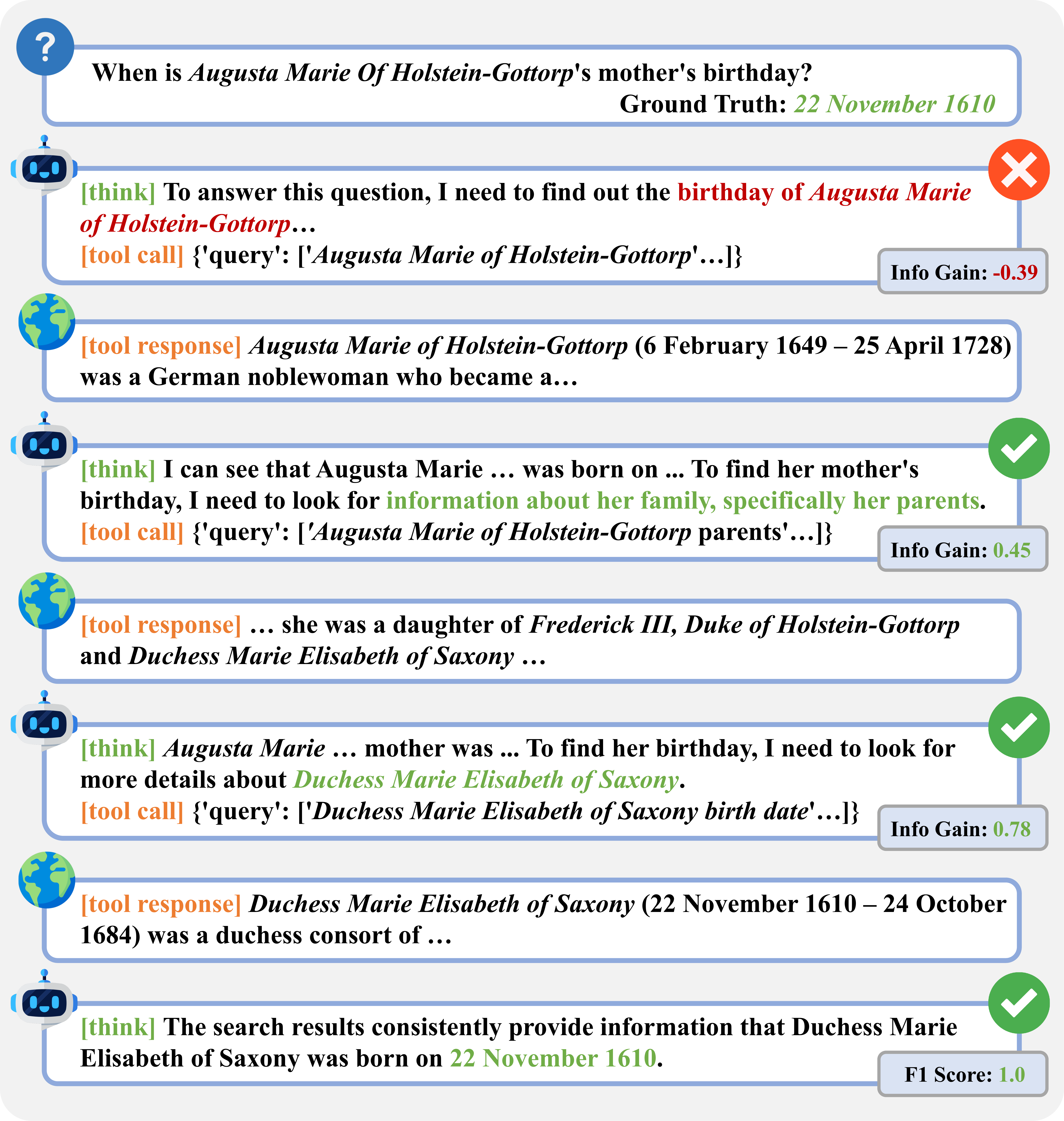}
    \caption{Case study illustrating a situation where the first round of retrieval failed, but the second and third rounds successfully located the correct evidence and produced the right answer. In this case, IGPO imposes a penalty on the erroneous retrieval in the first round.}
    \label{case2}
\end{figure}

\end{document}